%% file: main_eccv.tex
\documentclass[runningheads]{llncs}

\usepackage{graphicx}
\usepackage{amsmath,amssymb} 
\usepackage{color}
\usepackage{epsfig}
\usepackage{comment}
\usepackage{kevinlin}
\usepackage[ruled,vlined]{algorithm2e}
\usepackage{enumitem}
\usepackage[export]{adjustbox}
\usepackage{tabularx}
\usepackage[table]{xcolor}
\usepackage{colortbl}
\usepackage{subfig}
\usepackage{multirow}
\usepackage{booktabs}
\usepackage{xr}
\usepackage{breakcites}

\usepackage[width=122mm,left=12mm,paperwidth=146mm,height=193mm,top=12mm,paperheight=217mm]{geometry}

\begin{document}

\pagestyle{headings}
\mainmatter

\title{Adversarial Learning for Fine-grained Image Search} 

\titlerunning{Adversarial Learning for Fine-grained Image Search}

\authorrunning{K. Lin et al.}


\author{Kevin Lin$^1$, Fan Yang$^2$, Qiaosong Wang$^2$, Robinson Piramuthu$^2$}
\institute{$^1$University of Washington $^2$eBay Inc.\\
\email{kvlin@uw.edu, \{fyang4, qiaowang, rpiramuthu\}@ebay.com}
}

\maketitle

\begin{abstract}
Fine-grained image search is still a challenging problem due to the difficulty in capturing subtle differences regardless of pose variations of objects from fine-grained categories. In practice, a dynamic inventory with new fine-grained categories adds another dimension to this challenge.
In this work, we propose an end-to-end network, called FGGAN, that learns discriminative representations by implicitly learning a geometric transformation from multi-view images for fine-grained image search. 
We integrate a generative adversarial network (GAN) that can automatically handle complex view and pose variations by converting them to a canonical view without any predefined transformations. 
Moreover, in an open-set scenario, our network is able to better match  images from unseen and unknown fine-grained categories.
Extensive experiments on two public datasets and a newly collected dataset have demonstrated the outstanding robust performance of the proposed FGGAN in both closed-set and open-set scenarios, providing as much as 10\% relative improvement compared to baselines.

\end{abstract}

\input{01-intro}
\input{02-related-works}
\input{03-approach}
\input{04-exp}

\input{05-conclusion}

\bibliographystyle{splncs}
\bibliography{bib-gan,bib-retrieval,bib-finegrained,bib-geometric.bib}

\end{document}

%% file: 01-intro.tex
\section{Introduction}\label{sec:intro}

While image search has been extensively studied, it still remains a challenging problem~\cite{DBLP:conf/iccv/SivicZ03,DBLP:conf/mm/WanWHWZZL14,DBLP:conf/kdd/JingLKZXDT15,DBLP:conf/kdd/YangKBSWKP17}. In particular, it is extremely difficult to identify images at a fine-grained level, where the goal is to find objects belonging to the same fine-grained category as the query, \eg, identifying the make and model of cars. 
Numerous algorithms using deep neural networks have achieved state-of-the-art performance on fine-grained categorization~\cite{DBLP:conf/cvpr/DengK013,DBLP:conf/icpr/KrauseGDLF14,DBLP:conf/cvpr/KrauseJYL15,DBLP:conf/cvpr/QianJZL15,DBLP:conf/cvpr/XieYWL15,lin2015bilinear,DBLP:conf/cvpr/CuiZLB16,zhou2016fine,cui2017kernel}, but they are not directly applicable to fine-grained image search. 
Practically, given a dynamic inventory in production, the image search system needs to be sufficiently robust when new products are included.
While fine-grained \emph{categorization} mainly operates on a closed dataset containing a fixed number of categories, it could not handle unseen categories well.
Although classifiers can be re-trained to accommodate new categories, frequent re-training becomes prohibitively expensive as new data accumulates.
In contrast, fine-grained \emph{image search} by design should be aware of unseen categories that are not part of the training set~\cite{DBLP:journals/corr/YaoZZLT17a}. 

\setlength{\textfloatsep}{12pt}
\begin{figure}[t]
\begin{center}
\includegraphics[width=.7\linewidth]{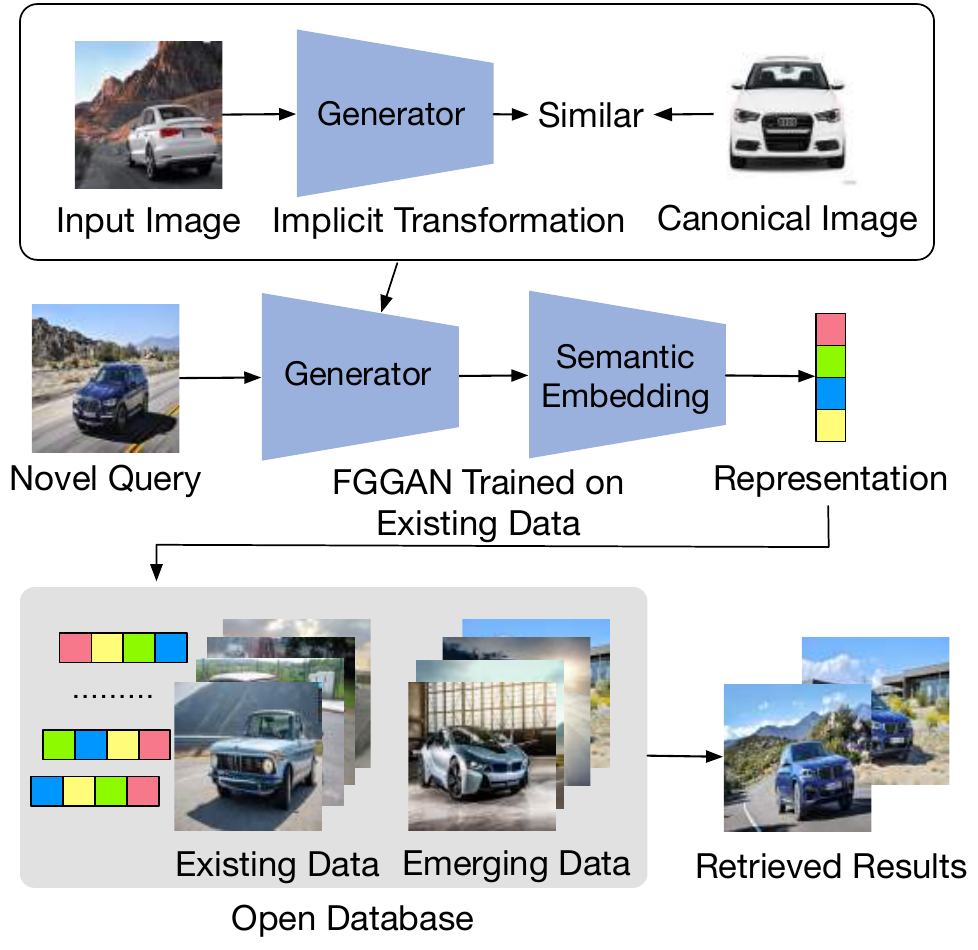}
\end{center} 
\vspace{-3mm}
\caption{The main idea of the proposed approach. 
We propose to generate image representations with adversarial networks by learning implicit transformations to normalize view and pose. The generated representation generalizes well for fine-grained image search given unseen categories.}
\label{fig:idea}
\end{figure}

In addition to emerging categories, view and pose variations of objects make finding correct fine-grained categories even harder. 
Classic approaches to address pose variations rely on matching local feature points, refining the homography, and inferring an explicit geometric transformation~\cite{DBLP:conf/cvpr/PhilbinCISZ07,DBLP:journals/ijcv/JegouDS10,DBLP:conf/cvpr/LiLH15}, but they are computationally expensive.
Recent works based on deep neural networks introduce dedicated modules to learn specific geometric transformations for semantic correspondence~\cite{DBLP:conf/nips/JaderbergSZK15,DBLP:conf/nips/ChoyGSC16,DBLP:conf/cvpr/KanazawaJC16,DBLP:journals/corr/RoccoAS17,DBLP:journals/corr/HanRHWCSP17,DBLP:journals/corr/KimMHJLS17}, of which representative works include spatial transformer network~\cite{DBLP:conf/nips/JaderbergSZK15}, universal correspondence network~\cite{DBLP:conf/nips/ChoyGSC16} and WarpNet~\cite{DBLP:conf/cvpr/KanazawaJC16}, \etc. 
Nevertheless, they require a pre-defined transformation type and a well-initialized transformation matrix beforehand to ensure reasonable performance, and cannot handle complex transformations. Therefore, they are impractical for fine-grained image search given a growing database that contains unknown transformations.

To address such problems, we resort to the generative adversarial network (GAN)~\cite{goodfellow2014generative} that shows outstanding performance on generating highly realistic images for various tasks. In this work, rather than generating high-quality images, we integrate GAN into a multi-task network to rectify view and pose variations, and jointly learn discriminative features for search (see Figure~\ref{fig:idea}). 
Specifically, the proposed network, called FGGAN, consists of two main components: a \emph{generator} and an \emph{evaluator}. While the generator is a fully convolutional network, the evaluator is composed of three sub-modules: a \emph{discriminator}, a \emph{normalizer} and a \emph{semantic embedding}. 
The generator and discriminator combined are analogous to the architecture of classic GANs. 
The normalizer learns implicit and class-agnostic geometric transformations to normalize an object in various views/poses to a canonical view without any pre-defined transformation parameters. 
The semantic embedding module enforces images from the same fine-grained category to have similar feature representations that are further used for retrieval. 
The three sub-modules of the evaluator are jointly optimized together with the generator, so that they are balanced to contribute to a good representation.
Our network removes the hassle of explicitly learning a geometric transformation and enables end-to-end training and inference to match objects in various views and poses, thus more flexible for real applications.  

Our motivation lies in two aspects. On one hand, GANs capture underlying data distribution without human supervision.
By traversing on the manifold, we can freely manipulate images to recover different poses from a single pose.
On the other hand, GANs abstract specific patterns from training images that are generalizable to a broader range of categories. 
Therefore, given only a small amount of training data, we are able to identify and match objects from categories that the network has never seen before. 
To the best of our knowledge, this is the first attempt to apply GAN to fine-grained image search in an open-set scenario.

Our contribution can be summarized in four-fold. 1) We apply GAN to fine-grained image search by implicitly learning transformations to match objects in various views and poses. 
2) Our adversarial learning setting only requires a small number of training samples and generalizes well to unseen categories. 
3) Our approach consistently outperforms its counterparts on several datasets in both closed-set and open-set scenarios. 
4) We construct and will release a new dataset for fine-grained image search that may benefit future research. 

%% file: 02-related-works.tex
\section{Related works}\label{sec:related-works}

{\flushleft \textbf{Fine-grained categorization and retrieval}} are two related but different tasks. 
Fine-grained categorization has been extensively investigated and various algorithms have achieved outstanding performance~\cite{DBLP:conf/cvpr/DengK013,DBLP:conf/icpr/KrauseGDLF14,DBLP:conf/cvpr/KrauseJYL15,DBLP:conf/cvpr/QianJZL15,DBLP:conf/cvpr/XieYWL15,lin2015bilinear,DBLP:conf/cvpr/CuiZLB16,zhou2016fine,cui2017kernel}.
In contrast, fine-grained image search is under-studied and requires distinguishing subtle differences given a potentially growing database. 
Some works focus on sketch-based image search for product matching~\cite{DBLP:conf/sigir/HuangCJZZ17,Song_2017_ICCV}. 
Regarding natural images, Xie~\etal~\cite{DBLP:journals/tmm/XieWZT15} associated visual words with fine-grained semantic attributes learned from hand-crafted features and built a hierarchical database for evaluation. 
Recently, Wei~\etal~\cite{DBLP:journals/tip/WeiLWZ17} proposed to selectively aggregate convolutional descriptors to obtain more discriminative features. 
While they are evaluated on a closed dataset, Yao~\etal~\cite{DBLP:journals/corr/YaoZZLT17a} designed a one-shot learning strategy to identify unseen objects by learning with an incomplete auxiliary training set, which is similar to our problem. However, it requires multiple networks with several post-processing steps to achieve good performance, thus not scalable enough.


{\flushleft \textbf{Generative adversarial networks}}, recently proposed by Goodfellow~\etal~\cite{goodfellow2014generative}, show promising results in generating realistic images from random signals~\cite{radford2015unsupervised,DBLP:journals/corr/ZhangXLZHWM16,DBLP:journals/corr/NguyenYBDC16,DBLP:conf/eccv/ZhuKSE16,DBLP:conf/icml/ReedAYLSL16,DBLP:journals/corr/YangKBP17}. 
It has been applied to various problems including style transfer~\cite{DBLP:conf/eccv/YooKPPK16,pix2pix2016,DBLP:journals/corr/ZhuPIE17,DBLP:journals/corr/LiuBK17}, image super-resolution~\cite{ledig2016photo,DBLP:journals/corr/Wu0ZH17} and face editing~\cite{DBLP:journals/corr/ShuYHSSS17}, \etc. 
Very recently, a few works have applied GAN to a retrieval network to facilitate learning better representations. 
Wang~\etal~\cite{DBLP:conf/sigir/WangYZGXWZZ17} proposed IRGAN that iteratively optimizes a discriminative model and a generative model for information retrieval, where the two models compete with each other in an adversarial way.
Zheng~\etal~\cite{DBLP:journals/corr/ZhengZY17} applied GAN to generate high-quality images for training data augmentation in a semi-supervised manner for person re-identification.
Qiu~\etal~\cite{DBLP:conf/sigir/QiuPYM17} and Song~\cite{DBLP:journals/corr/abs-1708-04150} included GAN as an additional module to help learn compact binary features for fast retrieval. 
Creswell~\etal~\cite{DBLP:conf/eccv/CreswellB16} targeted on sketch retrieval by fine-tuning a conventional GAN and using the discriminator without the final layer as an encoder for feature embedding.
Our work differs from these methods as we use GAN to learn an implicit transformation to rectify view and pose variations for both closed-set and open-set scenarios, rather than generating images to augment training data. 

%% file: 03-approach.tex
\section{Approach}\label{sec:approach}

\begin{figure}[t]
	\centering
	\includegraphics[width=0.7\columnwidth]{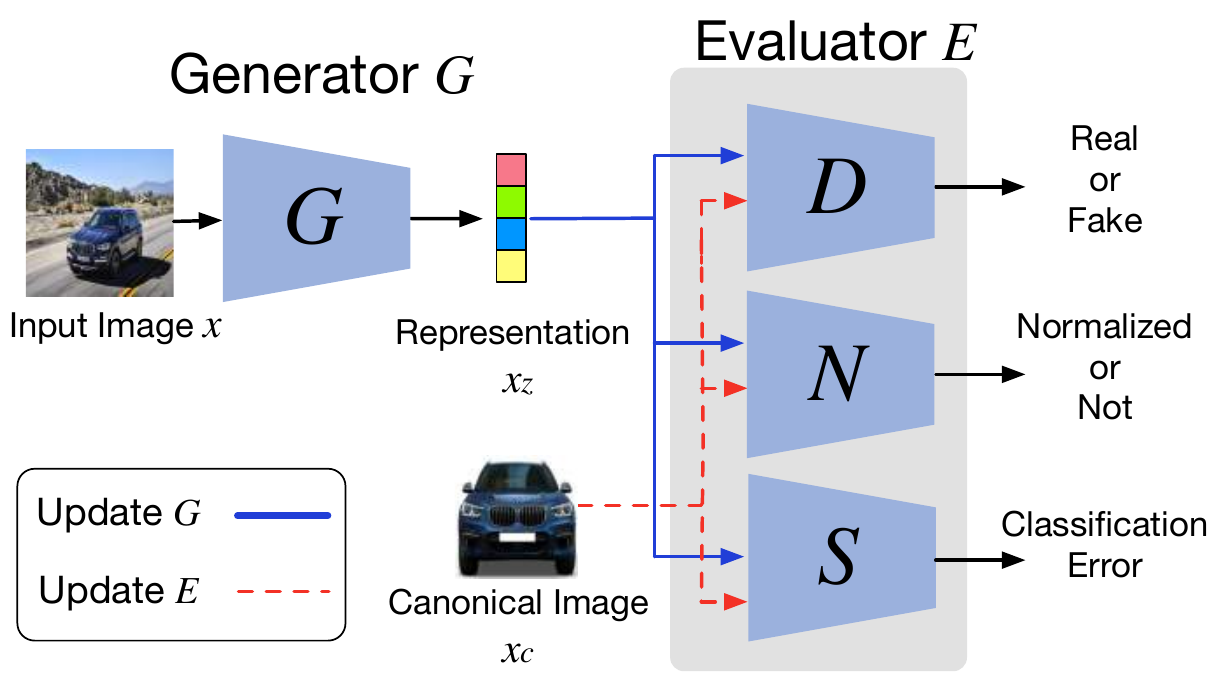}
	\caption{The architecture of the proposed FGGAN to learn discriminative features for fine-grained image retrieval. $D$, $N$ and $S$ denote discriminator,  normalizer and  semantic embedding, respectively. Alternating optimization is used to train $G$ and $E$.
}
	\label{fig:overview}
\end{figure}


\subsection{Overview}
Figure~\ref{fig:overview} illustrates the overview of our network architecture. 
Specifically, the proposed FGGAN consists of a generator $G$ and an evaluator $E$. 
The generator $G$ is trained to confuse the evaluator $E$ by producing high-quality features, while the evaluator $E$ aims at distinguishing the features generated by the generator $G$ from the real ones by optimizing multiple learning goals. This process can be formulated as
\begin{equation}
    \min_{\theta} \max_{\phi} \mathcal{L}(G, E)
    \label{eq:gan_form}
\end{equation}
where $\theta$ and $\phi$ denote the parameters of the generator $G$ and the evaluator $E$, respectively.
Given a real input image $x_r$ in any views or poses from the fine-grained category $y$, a new representation $x_z$ in feature space is obtained from the generator function $G(x_r)$.
In addition, given another input image $x_c$ in the canonical view from the same category, the evaluator $E$ evaluates the quality of $x_z$ via the following objective function:
\begin{equation}
    \mathop{\mathbb{E}}_{x_r \sim \mathcal{P}_{data}}\left[ \log E(x_r, x_c) \right] + \mathop{\mathbb{E}}_{x_z \sim \mathcal{P}_{G(x_r)}} \left[\log(1-E(x_z,x_c)) \right]
    \label{eq:e_loss}
\end{equation}
where $\mathcal{P}_{data}$ denotes the data distribution of real images in random views, and $\mathcal{P}_{G(x_r)}$ denotes the distribution of generated image representations given $x_r$.


To jointly learn implicit geometric transformations and generate discriminative image representations, the evaluator $E$ consists of three sub-modules: the discriminator $D$, the normalizer $N$ and the semantic embedding module $S$. 
The learning objective of $E$ is then written as
\begin{equation}
    E(x, x_c) = \gamma_D \mathcal{L}_D(x)+\gamma_N \mathcal{L}_N(x, x_c)+ \gamma_S\mathcal{L}_S(x)
    \label{eq:evaluator}
\end{equation}
where $x$ can be either the real image $x_r$ or the generated representation $x_z$. Additionally, $\gamma_D$, $\gamma_N$ and $\gamma_S$ are the hyper-parameters balancing the effect of each objective and they are set to equal value.


We describe the intuition behind each objective function as below. 
First, $\mathcal{L}_D(x)$ is defined as a binary classification loss function for the discriminator $D$ to classify the input into real and fake classes. Second, $\mathcal{L}_N(x,x_c)$ verifies whether the input is normalized to the canonical view/pose by computing the normalization error given the input pair $\{x,x_c\}$. Finally, $\mathcal{L}_S(x)$ encourages images from the same fine-grained category to have similar representations. We elaborate on each objective function as below.

\subsection{Discriminator}\label{sec:disc}
In the proposed FGGAN, the output of the discriminator \textcolor{black}{$D(x)$} is a scalar probability indicating whether the given input \textcolor{black}{$x$} is the real image or the generated image representation. The higher the probability \textcolor{black}{$D(x)$} is, the more chance $x$ would be the real image.
Following the definition in~\cite{goodfellow2014generative}, we formulate the following binary cross-entropy loss function to learn the discriminator $D$:
\begin{equation}
    \mathcal{L}_D(\textcolor{black}{x}) = 
        \begin{cases}
      -\log D(\textcolor{black}{x}), & \textcolor{black}{x} \sim \mathcal{P}_{data} \\
      -\log(1-D(\textcolor{black}{x})), & \textcolor{black}{x} \sim \mathcal{P}_{G(x)}
    \end{cases}
    \label{eq:d}
\end{equation}
The objective function tries to distinguish generated image representations from real images by alternating learning goals between the two cases.

\subsection{Normalizer}
One of the challenges in fine-grained image search is that objects in images may appear in high variation of viewpoints and poses. 
Different from previous works~\cite{DBLP:conf/cvpr/KanazawaJC16,DBLP:journals/corr/HanRHWCSP17,zhang2014part,zhang2013deformable,chai2013symbiotic} that localize part regions or match objects with pre-defined geometric transformations, we propose to implicitly learn the transformation by an end-to-end adversarial network, which normalizes various views and poses to a single view for better matching. 
While the generator function $G(x_r)$ learns to convert the input image $x_r$ to $x_z$ in the canonical view, we design a normalizer $N$ to distinguish the real canonical image $x_c$ from the generated pose-normalized representation $x_z$. 

Given a pair of input $\{x,x_c\}$, we define it as the positive pair when $x$ is the canonical image $x_c$ and as the negative pair if $x$ is the generated representation $x_z$. We concatenate the inputs $x$ and $x_c$ along the channel, forming a tensor as input for the normalizer $N$. Then, we train the normalizer $N$ as a binary classifier identifying whether the input pair is the positive or negative pair. In this way, the normalizer verifies whether the input $x$ is similar to $x_c$, \ie, how well $x$ is normalized to the canonical view. 
Accordingly, we train the generator to confuse the normalizer by generating image representation $x_z$ that is similar to the canonical image $x_c$, without any pre-defined transformation parameters. 
We formulate the normalization loss function as below:
\begin{equation}
	\begin{aligned}
    {\small 
    \mathcal{L}_N(x,x_{c}) = 
    \begin{cases}
    -\log N(x,x_{c}) - \lambda \ell(x, x_c), & x=x_c \\
    -\log(1-N(x,x_{c})) - \lambda \ell(x, x_c), & x \sim \mathcal{P}_{G(x)}
    \end{cases}
    }
    \label{eq:n}
    \end{aligned}
\end{equation}
To encourage that the generated representation is close to the real canonical image in feature space rather than forcing them to match exactly, we add a feature reconstruction loss $\ell(x, x_c) = \frac{1}{2}\|\mathbf{f}_x - \mathbf{f}_{x_{c}}\|^2$. 
Specifically, given the input pair $\{x, x_c\}$, we vectorize the activations of the intermediate layer of the normalizer to obtain two feature vectors $\mathbf{f}_x$ and $\mathbf{f}_{x_c}$. We minimize the mean square error (MSE) of $\mathbf{f}_x$ and $\mathbf{f}_{x_c}$ during learning. The weighting parameter $\lambda=1$.

\subsection{Semantic embedding}

To ensure that features of images from the same fine-grained category are semantically close to each other, we include a convolutional neural network with a classification loss to learn discriminative image representations while preserving semantic similarity.
\textcolor{black}{The semantic embedding module $S$ is able to evaluate the quality of the generated representations by estimating the classification error. Particularly, we firstly train $S$ with real images to capture the semantics in the feature space. Then, jointly learning with the generator $G$, we feed the generated representation $x_z$ into $S$ and compute the classification error. We then back-propagate the classification error into $G$ to help $G$ learn better.
} 
The objective function $\mathcal{L}_S$ is defined as a softmax loss: 
\begin{equation}
    \mathcal{L}_S(x) = -\log \frac{\exp(p_y(x))}{\sum_{i=1}^{C} \exp(p_i(x))}
    \label{eq:s}
\end{equation}
where $x$ can be the real image $x_r$ or the generated representation $x_z$. In addition, $y$ is the category label, $p_i(x)$ indicates the prediction score of the $i$-th category given the input $x$, and $C$ denotes the total number of categories.

\subsection{Adversarial training}
{\flushleft \textbf{Initialization.}} We define $\theta$, $\phi_D$, $\phi_N$, and $\phi_S$ as the parameters of the generator $G$, the discriminator $D$, the normalizer $N$ and the semantic embedding $S$, respectively. In the initialization stage, the network parameters $\theta$, $\phi_D$, and $\phi_N$ are randomly initialized. $\phi_S$ is initialized with the weights pre-trained on ILSVRC12. Following the adversarial training procedure in~\cite{goodfellow2014generative}, we train $G$ and $E$ alternatively as below. 
{\flushleft \textbf{Evaluator.}} First, we train $D$ with real images and generated representations by minimizing the loss of Eq.~\eqref{eq:d}. Second, $N$ is trained with the positive and negative pairs by minimizing the loss of Eq.~\eqref{eq:n}. 
$S$ is trained by optimizing the objective function Eq.~\eqref{eq:s}.
Finally, we freeze the parameters of the evaluator $E$, and then train the generator $G$.
{\flushleft \textbf{Generator.}} We optimize the parameters of the generator to \textit{increase} the loss of the evaluator, \ie, we optimize $\mathcal{L}_G(x,x_c) = -E(x, x_c)$ for the generator $G$.
Then, we freeze the parameters of the generator $G$, and then train the evaluator $E$. The training procedure will continue alternatively until the generator $G$ and the evaluator $E$ reach an equilibrium. Readers may refer to \textbf{Algorithm 1} for further details of the training procedure. 
\textcolor{black}{We implement our approach by using open source Torch~\cite{collobert:2011c} with multiple NVIDIA Telsa K80 GPUs. Following~\cite{radford2015unsupervised}, we train FGGAN with the stochastic gradient decent (SGD) with learning rate $0.0002$, mini-batch size $32$, and momentum $0.5$. We set the leak slope of LeakyReLU to $0.2$.
We multiply the classification error of the semantic embedding module $S$ by a factor of $0.001$ to facilitate training with other modules.
Further details of the network configuration can be found in the supplementary material.}
\newlength{\textfloatsepsave} \setlength{\textfloatsepsave}{\textfloatsep} \setlength{\textfloatsep}{0pt}
\begin{algorithm}[t]
   \label{agl:algo}
  \caption{Adversarial training of FGGAN}
  \KwIn{$N$ pairs of training data $\{x_r^i,x_c^i\}_{i=1}^N$ and corresponding category labels $\{y^i\}_{i=1}^N$; learning rate $\alpha$ and batch size $\beta$.}
  \KwOut{Network parameters $\theta$,~$\phi_D$,~$\phi_N$,~$\phi_S$}
  \textbf{Initialization:} 
  Initialize parameters~$\theta$,~$\phi_D$,~$\phi_N$,~$\phi_S$\\
  \textbf{Optimization:}\\
  \While{not reach an equilibrium}
  {
  \BlankLine
  Sample $x^i$ uniformly from $\{x_r^i, G(x_r^i)\}$\\
	Update the network parameters $\phi_D$:\\
    $\phi_D \leftarrow \phi_D - \alpha \frac{1}{\beta}\sum_{i=1}^{\beta}\frac{\partial \mathcal{L}_D(x^i)}{\partial \phi_D}$\\   
    Update the network parameters $\phi_N$:\\
    $\phi_N \leftarrow \phi_N - \alpha \frac{1}{\beta}\sum_{i=1}^{\beta}\frac{\partial \mathcal{L}_N(x^i,x_c^i)}{\partial \phi_N}$\\ 
    Update the network parameters $\phi_S$:\\
    $\phi_S \leftarrow \phi_S - \alpha \frac{1}{\beta}\sum_{i=1}^{\beta}\frac{\partial \mathcal{L}_S(x^i)}{\partial \phi_S}$\\     
    Update the network parameters $\theta$:\\
    $\theta \leftarrow \theta - \alpha \frac{1}{\beta}\sum_{i=1}^{\beta}\frac{\partial \mathcal{L}_G(x^i,x_c^i)}{\partial \theta}$ 
  }
\end{algorithm}

\subsection{Image retrieval}
After training, we feed an input image into $G$ and generate the representation in the canonical view (see Figure \ref{fig:idea}), which is further fed into the semantic embedding module $S$. The activations of the last fully-connected layer or compact representations aggregated from the activations of the last pooling layer of $S$ are extracted as feature vectors for image retrieval using Euclidean distance. We evaluate both in the experiments.

\setlength{\textfloatsep}{\textfloatsepsave}

%% file: 04-exp.tex
\section{Experiments}\label{sec:exp}
\subsection{Evaluation protocol}
We adopt \textbf{precision at $k$~(P@$k$)} and \textbf{mean average precision~(mAP)} as evaluation metrics.
P@$k$ indicates the percentage of true positive samples among the top $k$ returned samples. The mAP indicates the mean of the average precision scores given a set of queries.
Following the common protocol, we select query images from the test set, and retrieve nearest neighbors from the training set. 
We take the fine-grained class labels as ground truth to compute the metrics by validating whether the query and retrieved images share common labels.

\begin{table*}[t]
\renewcommand{\arraystretch}{0.95}
	\centering
    \small
	\caption{Mean average precision (mAP, \%) of different methods evaluated on top $k$ retrieved images on CompCars, eBayCamera10k, and Lookbook datasets.
    Our full model is better than or comparable with other methods.
    \emph{ft} denotes the backbone network (VGG16) fine-tuned on the training set of the target dataset.}
	\label{tbl:map-compare}
    {
	\begin{tabular*}{\textwidth}{@{}@{\extracolsep{\fill}}lcccc|cccc|ccc@{}}
	& \multicolumn{3}{c}{CompCars} && \multicolumn{3}{c}{eBayCamera10k} && \multicolumn{3}{c}{Lookbook} \\
	\hline
	Method  & $k = 5$ &  $k = 10$ & $k = 20$ && $k = 5$ &  $k = 10$ & $k = 20$ && $k = 5$ & $k = 10$ & $k = 20$\\
	\hline
    VGG16~\cite{Simonyan14c} & $47.32$ & $45.04$ & $40.56$ && $78.66$ & $75.21$ & $72.17$ && $52.52$ & $50.52$ & $46.93$\\  
        VGG16+\emph{ft}~\cite{Simonyan14c} & $63.18$ & $61.64$ & $55.09$ && $84.32$ & $81.45$ & $78.65$ && $76.98$ & $72.60$ & $67.64$\\
        \hline

MAC+\emph{ft}~\cite{tolias2015particular,azizpour2015generic} & $67.81$ & $63.41$ & $57.08$ && $86.00$ & $82.56$ & $79.42$ && $79.36$ & $74.79$ & $70.10$\\ 
    Sum pooling+\emph{ft}~\cite{babenko2015aggregating} & $66.78$ & $63.10$ & $56.91$ && $83.75$ & $80.60$ & $77.52$ && $80.53$ & $\textbf{76.31}$ & $71.65$\\ 
    CroW+\emph{ft}~\cite{kalantidis2016cross} & $66.22$ & $62.28$ & $56.11$ && $84.93$ & $81.87$ & $78.76$ && $80.48$ & $76.12$ & $\textbf{72.44}$\\  
 SCDA+\emph{ft}~\cite{DBLP:journals/tip/WeiLWZ17} & $65.69$ & $61.71$ & $55.75$ && $85.22$ & $82.20$ & $79.33$ && $66.03$ & $62.25$ & $58.19$\\
    \hline
    FGGAN & $66.47$ & $62.52$ & $56.28$ && $92.16$ & $\textbf{90.34}$ & $\textbf{87.81}$ && $77.00$ & $71.90$ & $66.98$\\
    FGGAN+MAC & $\textbf{67.86}$ & $\textbf{63.88}$ & $\textbf{57.28}$ && $92.03$ & $89.91$ & $87.70$ && $79.45$ & $74.94$ & $70.11$\\
    FGGAN+Sum pooling & $67.21$ & $63.07$ & $56.38$ && $91.34$ & $89.30$ & $86.86$ && $78.43$ & $74.00$ & $68.88$\\
    FGGAN+CroW & $66.58$ & $62.40$ & $56.28$ && $91.90$ & $89.93$ & $87.74$ && $\textbf{80.91}$ & $76.23$ & $71.38$\\
    FGGAN+SCDA & $66.43$ & $62.67$ & $56.16$ && $\textbf{92.22}$ & $90.05$ & $87.73$ && $71.96$ & $67.62$ & $63.12$\\
	\hline
	\end{tabular*}
    }
\end{table*}

\subsection{Dataset}

{\flushleft \textbf{CompCars~\cite{yang2015large}}} contains car images in various views including front, side, rear, front-side, and rear-side views, and rich annotations including make, model, and the year of manufacture. 
It is divided into three parts for fine-grained recognition, attribute prediction and car verification. 
Here we conduct experiments on the first part that consists of $30,730$ training images and $11,119$ test images from $431$ categories of cars. 
The year of car manufacture ranges from $2005$ to $2016$.

{\flushleft \textbf{Lookbook}~\cite{DBLP:conf/eccv/YooKPPK16}}~consists of two types of clothings images: street photos and stock images with clean background. 
Each category contains one stock image associated with multiple street photos.
Following~\cite{DBLP:conf/eccv/YooKPPK16}, we obtain $68,819$ street photos and $8,726$ stock images. 
We split them into training and test sets. The training set has $8,726$ stock images and the associated $39,201$ street photos, while the test set has $29,618$ street photos. This contains non-rigid objects.



{\flushleft \textbf{eBayCamera10k}} is a new fine-grained dataset compiled by us. It consists of $110$ fine-grained types of camera and lens in terms of make and model, including $471$ stock images and $10,720$ user photos downloaded from eBay.com.
Similar to Lookbook, the stock image and user photos are associated if both of them share a common make and model. 
The training set includes $471$ stock images and the associated $8,040$ user images. The test set has $2,680$ user images.
Cameras in the user photos may have high view and pose variations, or come with additional accessories, making accurate retrieval challenging. 
We will release the dataset to benefit future research.
\begin{figure*}[t]
 \centering
 \setlength{\tabcolsep}{0.4pt}
 \setlength{\fboxsep}{0pt}%
 \setlength{\fboxrule}{0.7pt}%
 \begin{tabular}{cccccccccccccccccccccccc}

 \includegraphics[width=.055\textwidth,height=.055\textwidth]{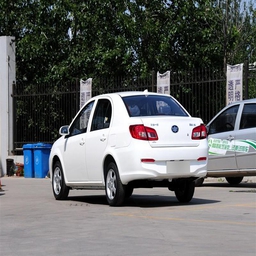}
 &
	\rotatebox[origin=l]{90}{\scalebox{0.5}{VGG$_{ft}$}}
    & \includegraphics[width=.055\textwidth,height=.055\textwidth,cfbox=cyan]{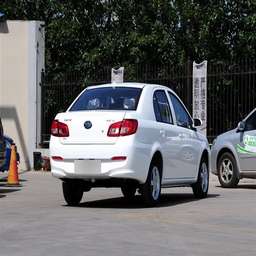}
    & \includegraphics[width=.055\textwidth,height=.055\textwidth,cfbox=cyan]{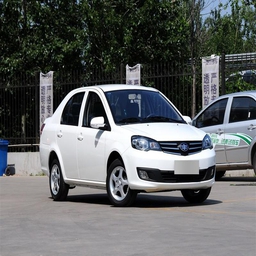}
    & \includegraphics[width=.055\textwidth,height=.055\textwidth]{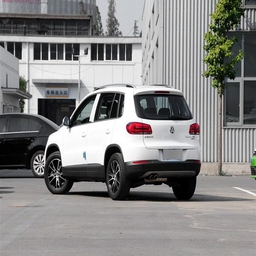}
    & \includegraphics[width=.055\textwidth,height=.055\textwidth]{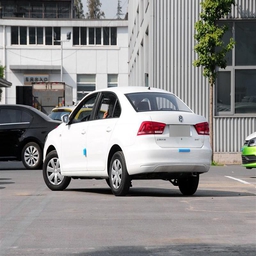}
    & \includegraphics[width=.055\textwidth,height=.055\textwidth]{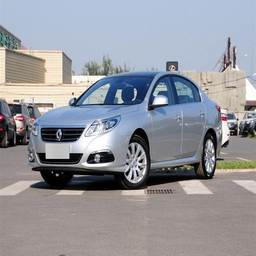}
    & ~~ 
    & \includegraphics[width=.055\textwidth,height=.055\textwidth]{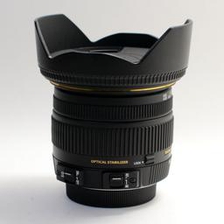}
	& \rotatebox[origin=l]{90}{\scalebox{0.5}{VGG$_{ft}$}}
    & \includegraphics[width=.055\textwidth,height=.055\textwidth,cfbox=cyan]{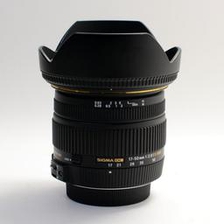}
    & \includegraphics[width=.055\textwidth,height=.055\textwidth]{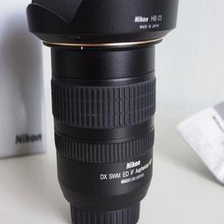}    & \includegraphics[width=.055\textwidth,height=.055\textwidth]{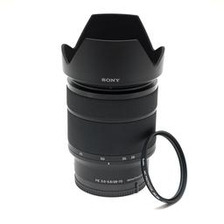}    & \includegraphics[width=.055\textwidth,height=.055\textwidth]{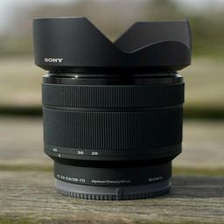}    & \includegraphics[width=.055\textwidth,height=.055\textwidth]{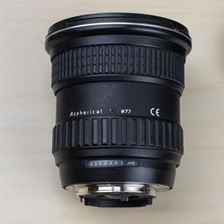}
    & ~~
    & \includegraphics[height=.055\textwidth]{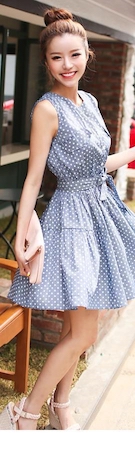}
	& \rotatebox[origin=l]{90}{\scalebox{0.5}{VGG$_{ft}$}}
    & \includegraphics[height=.055\textwidth,cfbox=cyan]{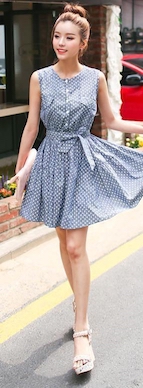}
    & \includegraphics[height=.055\textwidth]{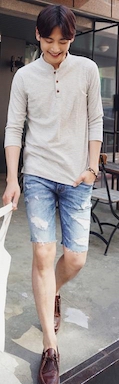}    & \includegraphics[height=.055\textwidth]{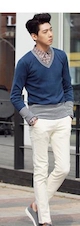}    & \includegraphics[height=.055\textwidth]{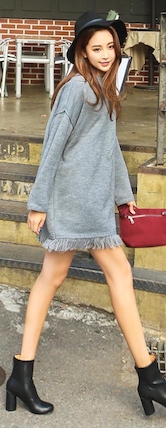}    & \includegraphics[height=.055\textwidth]{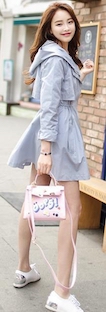}\\
    
    &
	\rotatebox[origin=l]{90}{\scalebox{0.5}{MAC$_{ft}$}}
    & \includegraphics[width=.055\textwidth,height=.055\textwidth,cfbox=cyan]{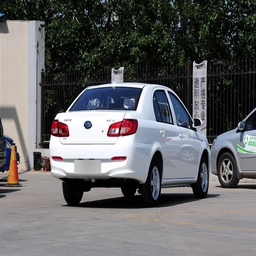}
    & \includegraphics[width=.055\textwidth,height=.055\textwidth]{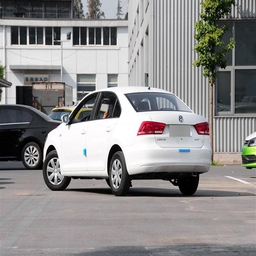}
    & \includegraphics[width=.055\textwidth,height=.055\textwidth]{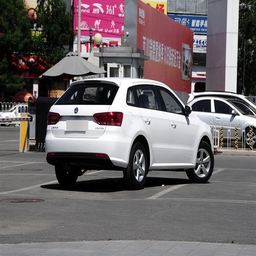}
    & \includegraphics[width=.055\textwidth,height=.055\textwidth]{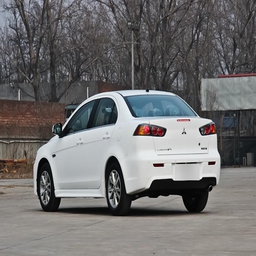}
    & \includegraphics[width=.055\textwidth,height=.055\textwidth]{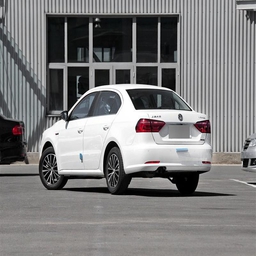}
    & ~~
    &
	& \rotatebox[origin=l]{90}{\scalebox{0.5}{MAC$_{ft}$}}
    & \includegraphics[width=.055\textwidth,height=.055\textwidth]{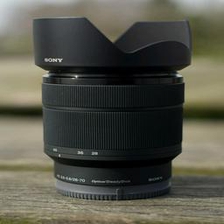}
    & \includegraphics[width=.055\textwidth,height=.055\textwidth]{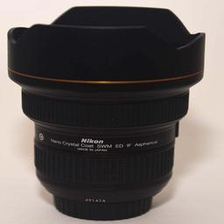}    & \includegraphics[width=.055\textwidth,height=.055\textwidth]{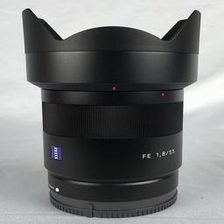}    & \includegraphics[width=.055\textwidth,height=.055\textwidth]{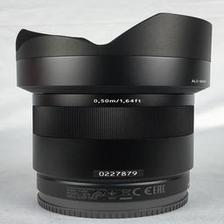}    & \includegraphics[width=.055\textwidth,height=.055\textwidth]{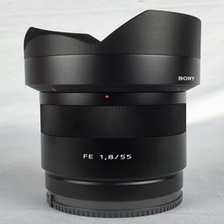}
    & ~~
    & 
	& \rotatebox[origin=l]{90}{\scalebox{0.5}{MAC$_{ft}$}}
    & \includegraphics[height=.055\textwidth,cfbox=cyan]{./figures/vis/lookbook/vggft/query900_rank_0_PID000109.jpg}
    & \includegraphics[height=.055\textwidth]{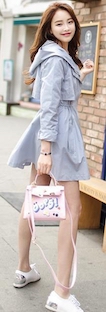}    & \includegraphics[height=.055\textwidth]{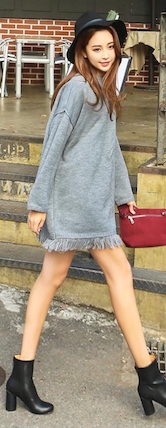}    & \includegraphics[height=.055\textwidth]{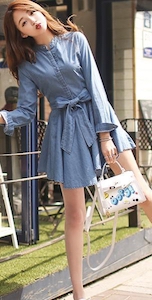}    & \includegraphics[height=.055\textwidth]{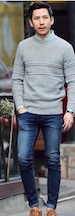}\\

    &
	\rotatebox[origin=l]{90}{\scalebox{0.5}{Sum$_{ft}$}}
    & \includegraphics[width=.055\textwidth,height=.055\textwidth,cfbox=cyan]{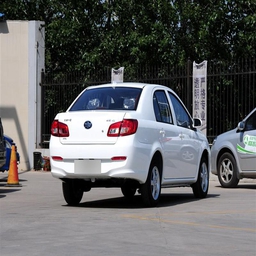}
    & \includegraphics[width=.055\textwidth,height=.055\textwidth]{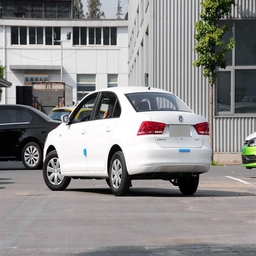}
    & \includegraphics[width=.055\textwidth,height=.055\textwidth,cfbox=cyan]{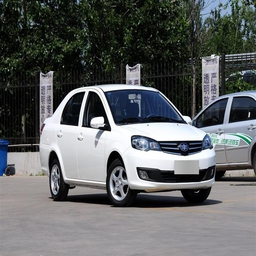}
    & \includegraphics[width=.055\textwidth,height=.055\textwidth]{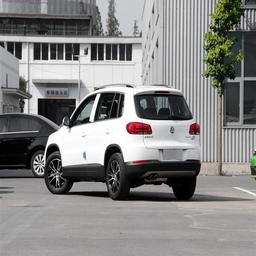}
    & \includegraphics[width=.055\textwidth,height=.055\textwidth]{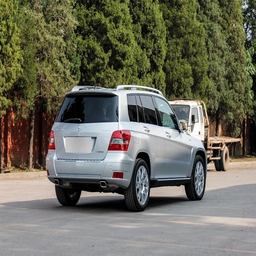}
    & ~~
    &
	& \rotatebox[origin=l]{90}{\scalebox{0.5}{Sum$_{ft}$}}
    & \includegraphics[width=.055\textwidth,height=.055\textwidth,cfbox=cyan]{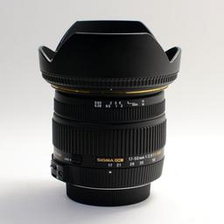}    & \includegraphics[width=.055\textwidth,height=.055\textwidth]{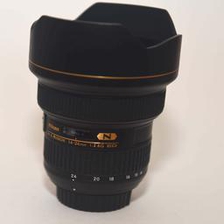}    & \includegraphics[width=.055\textwidth,height=.055\textwidth,cfbox=cyan]{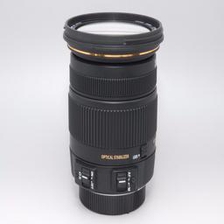}    & \includegraphics[width=.055\textwidth,height=.055\textwidth]{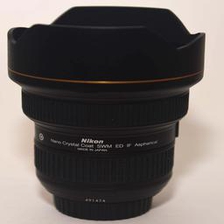}    & \includegraphics[width=.055\textwidth,height=.055\textwidth]{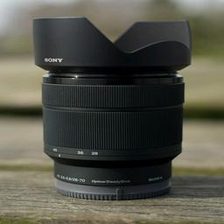}
    & ~~
    & 
	& \rotatebox[origin=l]{90}{\scalebox{0.5}{Sum$_{ft}$}}
    & \includegraphics[height=.055\textwidth,cfbox=cyan]{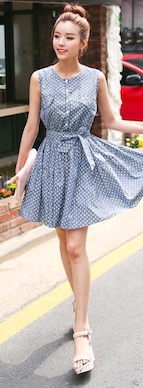}
    & \includegraphics[height=.055\textwidth,cfbox=cyan]{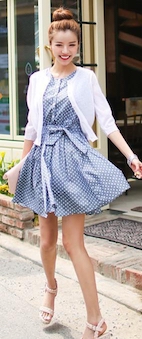}    & \includegraphics[height=.055\textwidth]{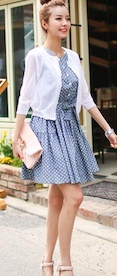}    & \includegraphics[height=.055\textwidth]{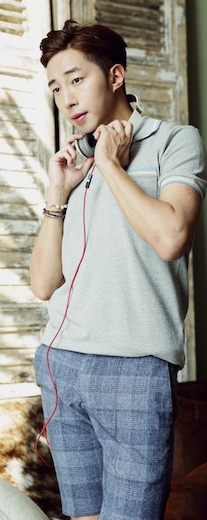}    & \includegraphics[height=.055\textwidth]{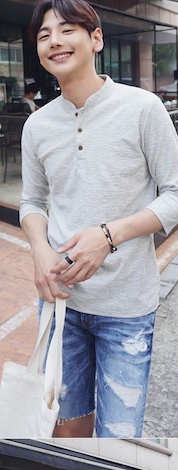}\\

    &
	\rotatebox[origin=l]{90}{\scalebox{0.5}{Crow$_{ft}$}}
    & \includegraphics[width=.055\textwidth,height=.055\textwidth,cfbox=cyan]{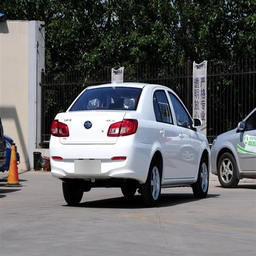}
    & \includegraphics[width=.055\textwidth,height=.055\textwidth]{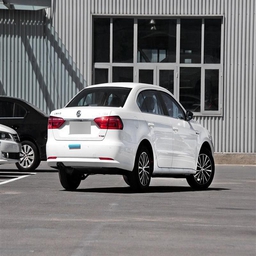}
    & \includegraphics[width=.055\textwidth,height=.055\textwidth]{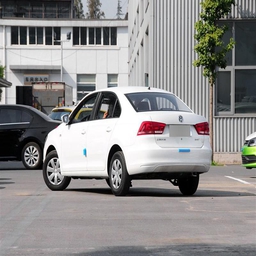}
    & \includegraphics[width=.055\textwidth,height=.055\textwidth,cfbox=cyan]{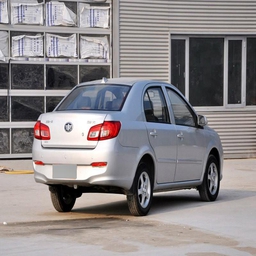}
    & \includegraphics[width=.055\textwidth,height=.055\textwidth]{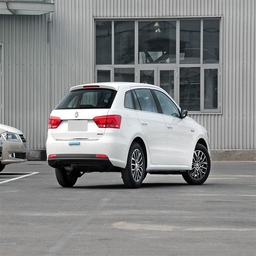}
    & ~~
    &
	& \rotatebox[origin=l]{90}{\scalebox{0.5}{Crow$_{ft}$}}
    & \includegraphics[width=.055\textwidth,height=.055\textwidth]{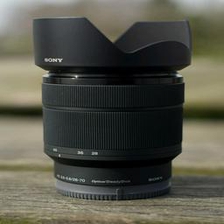}    & \includegraphics[width=.055\textwidth,height=.055\textwidth,cfbox=cyan]{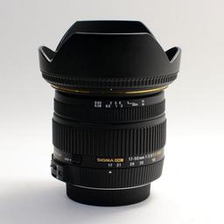}    & \includegraphics[width=.055\textwidth,height=.055\textwidth]{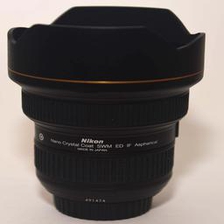}    & \includegraphics[width=.055\textwidth,height=.055\textwidth]{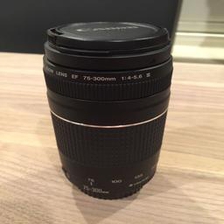}    & \includegraphics[width=.055\textwidth,height=.055\textwidth]{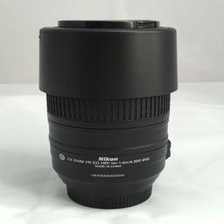}
    & ~~
    & 
	& \rotatebox[origin=l]{90}{\scalebox{0.5}{Crow$_{ft}$}}
    & \includegraphics[height=.055\textwidth,cfbox=cyan]{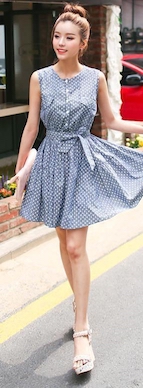}
    & \includegraphics[height=.055\textwidth]{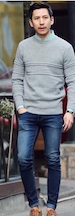}    & \includegraphics[height=.055\textwidth,cfbox=cyan]{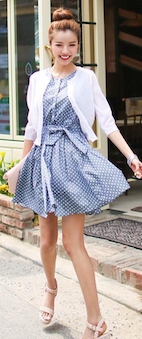}    & \includegraphics[height=.055\textwidth]{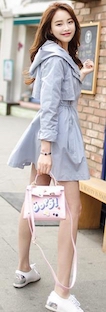}    & \includegraphics[height=.055\textwidth]{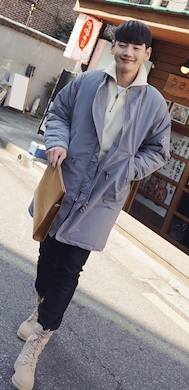}\\

    &
	\rotatebox[origin=l]{90}{\scalebox{0.5}{SCDA$_{ft}$}}
    & \includegraphics[width=.055\textwidth,height=.055\textwidth,cfbox=cyan]{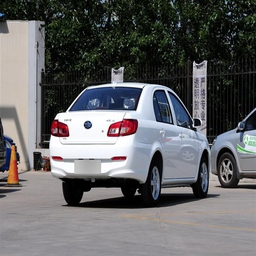}
    & \includegraphics[width=.055\textwidth,height=.055\textwidth]{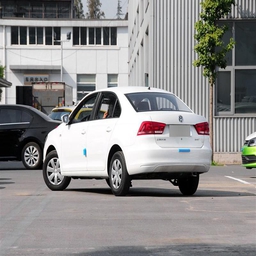}    & \includegraphics[width=.055\textwidth,height=.055\textwidth]{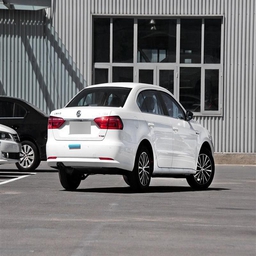}    & \includegraphics[width=.055\textwidth,height=.055\textwidth]{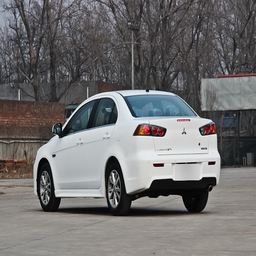}    & \includegraphics[width=.055\textwidth,height=.055\textwidth,cfbox=cyan]{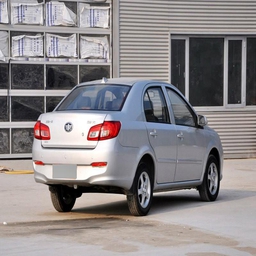}
    & ~~
    &
	& \rotatebox[origin=l]{90}{\scalebox{0.5}{SCDA$_{ft}$}}
    & \includegraphics[width=.055\textwidth,height=.055\textwidth]{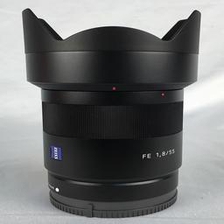}    & \includegraphics[width=.055\textwidth,height=.055\textwidth]{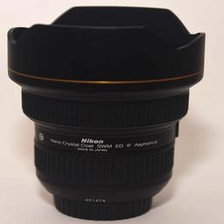}    & \includegraphics[width=.055\textwidth,height=.055\textwidth]{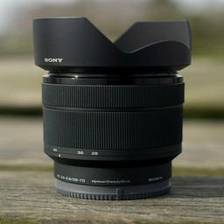}    & \includegraphics[width=.055\textwidth,height=.055\textwidth]{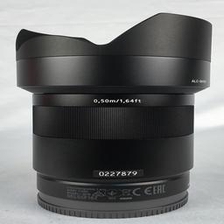}    & \includegraphics[width=.055\textwidth,height=.055\textwidth,cfbox=cyan]{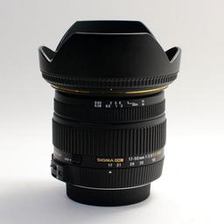}
    & ~~
    & 
	& \rotatebox[origin=l]{90}{\scalebox{0.5}{SCDA$_{ft}$}}
    & \includegraphics[height=.055\textwidth,cfbox=cyan]{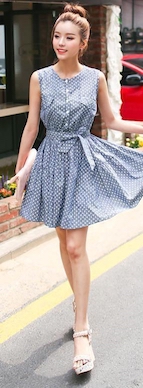}
    & \includegraphics[height=.055\textwidth]{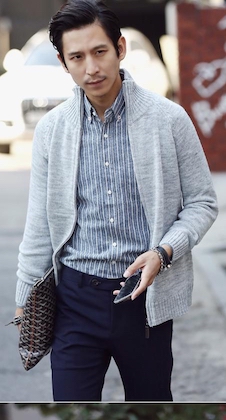}    & \includegraphics[height=.055\textwidth]{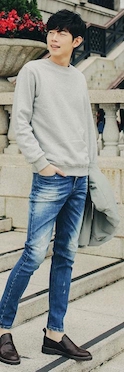}    & \includegraphics[height=.055\textwidth]{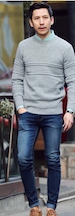}    & \includegraphics[height=.055\textwidth]{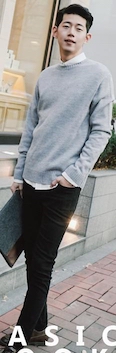}\\

    &
	\rotatebox[origin=l]{90}{\scalebox{0.5}{FGGAN}}
    & \includegraphics[width=.055\textwidth,height=.055\textwidth,cfbox=cyan]{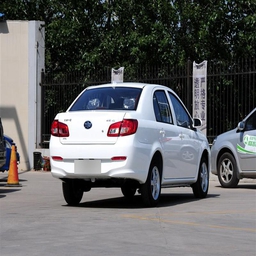}
    & \includegraphics[width=.055\textwidth,height=.055\textwidth]{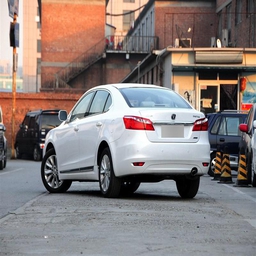}    & \includegraphics[width=.055\textwidth,height=.055\textwidth]{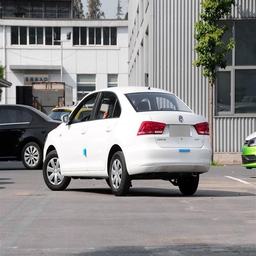}    & \includegraphics[width=.055\textwidth,height=.055\textwidth]{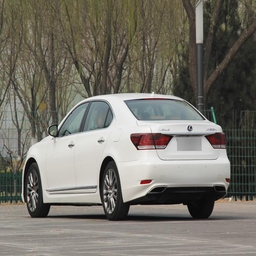}    & \includegraphics[width=.055\textwidth,height=.055\textwidth,cfbox=cyan]{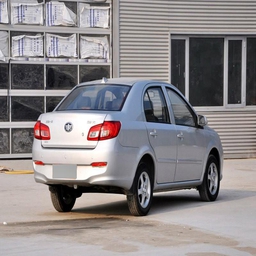}
    & ~~
    &
	& \rotatebox[origin=l]{90}{\scalebox{0.5}{FGGAN}}
    & \includegraphics[width=.055\textwidth,height=.055\textwidth,cfbox=cyan]{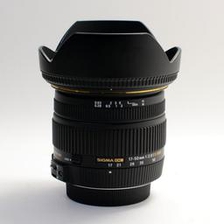}    & \includegraphics[width=.055\textwidth,height=.055\textwidth,cfbox=cyan]{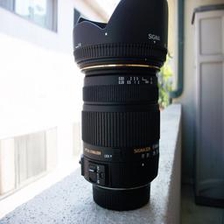}    & \includegraphics[width=.055\textwidth,height=.055\textwidth,cfbox=cyan]{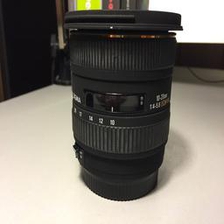}    & \includegraphics[width=.055\textwidth,height=.055\textwidth,cfbox=cyan]{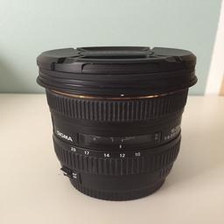}    & \includegraphics[width=.055\textwidth,height=.055\textwidth,cfbox=cyan]{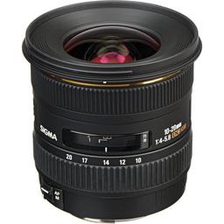}
    & ~~
    & 
	& \rotatebox[origin=l]{90}{\scalebox{0.5}{FGGAN}}
    & \includegraphics[height=.055\textwidth,cfbox=cyan]{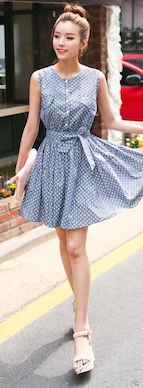}
    & \includegraphics[height=.055\textwidth]{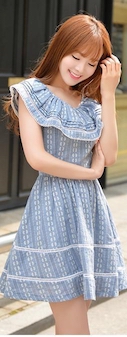}    & \includegraphics[height=.055\textwidth,cfbox=cyan]{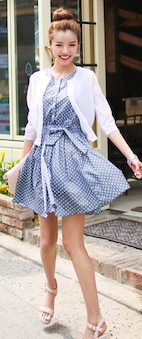}    & \includegraphics[height=.055\textwidth,cfbox=cyan]{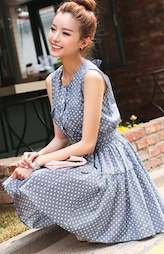}    & \includegraphics[height=.055\textwidth]{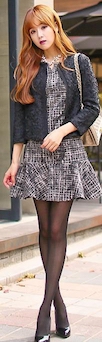}\\

    &
	\rotatebox[origin=l]{90}{\scalebox{0.5}{FG$_{MAC}$}}
    & \includegraphics[width=.055\textwidth,height=.055\textwidth,cfbox=cyan]{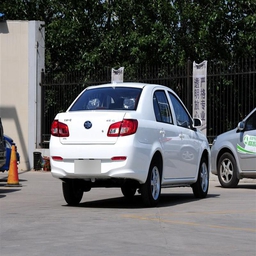}
    & \includegraphics[width=.055\textwidth,height=.055\textwidth]{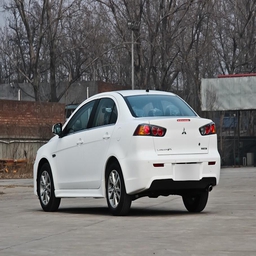}    & \includegraphics[width=.055\textwidth,height=.055\textwidth,cfbox=cyan]{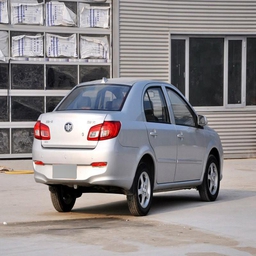}    & \includegraphics[width=.055\textwidth,height=.055\textwidth,cfbox=cyan]{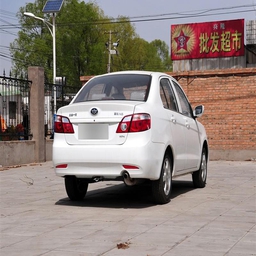}    & \includegraphics[width=.055\textwidth,height=.055\textwidth]{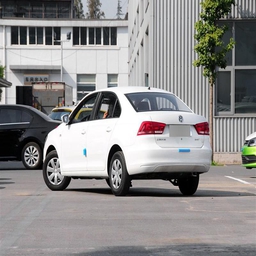}
    & ~~
    &
	& \rotatebox[origin=l]{90}{\scalebox{0.5}{FG$_{MAC}$}}
    & \includegraphics[width=.055\textwidth,height=.055\textwidth,cfbox=cyan]{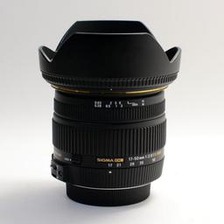}    & \includegraphics[width=.055\textwidth,height=.055\textwidth]{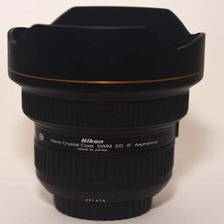}    & \includegraphics[width=.055\textwidth,height=.055\textwidth]{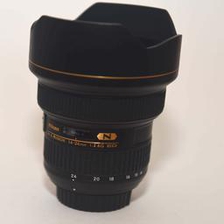}    & \includegraphics[width=.055\textwidth,height=.055\textwidth,cfbox=cyan]{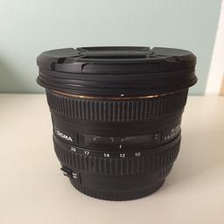}    & \includegraphics[width=.055\textwidth,height=.055\textwidth,cfbox=cyan]{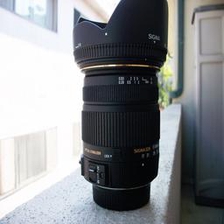}
    & ~~
    & 
	& \rotatebox[origin=l]{90}{\scalebox{0.5}{FG$_{MAC}$}}
    & \includegraphics[height=.055\textwidth,cfbox=cyan]{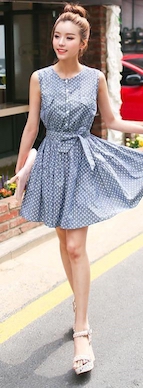}
    & \includegraphics[height=.055\textwidth]{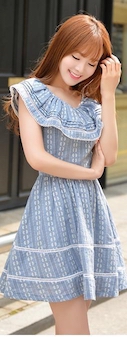}    
    & \includegraphics[height=.055\textwidth]{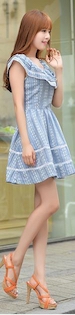}    
    & \includegraphics[height=.055\textwidth]{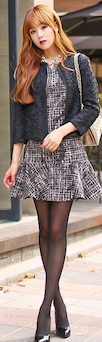}    
    & \includegraphics[height=.055\textwidth]{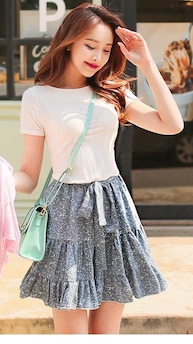}\\

    &
	\rotatebox[origin=l]{90}{\scalebox{0.5}{FG$_{Sum}$}}
    & \includegraphics[width=.055\textwidth,height=.055\textwidth,cfbox=cyan]{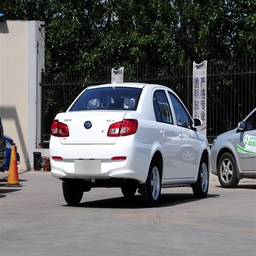}
    & \includegraphics[width=.055\textwidth,height=.055\textwidth,cfbox=cyan]{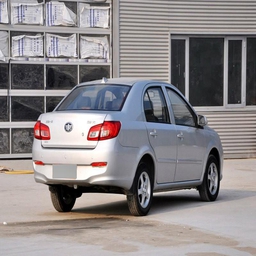}    & \includegraphics[width=.055\textwidth,height=.055\textwidth,cfbox=cyan]{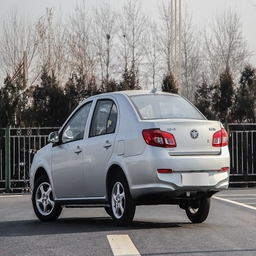}    & \includegraphics[width=.055\textwidth,height=.055\textwidth,cfbox=cyan]{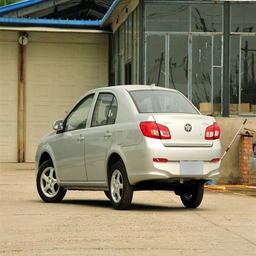}    & \includegraphics[width=.055\textwidth,height=.055\textwidth]{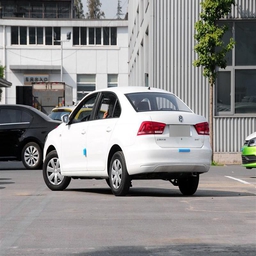}
    & ~~
    &
	& \rotatebox[origin=l]{90}{\scalebox{0.5}{FG$_{Sum}$}}
    & \includegraphics[width=.055\textwidth,height=.055\textwidth,cfbox=cyan]{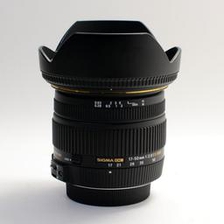}    & \includegraphics[width=.055\textwidth,height=.055\textwidth]{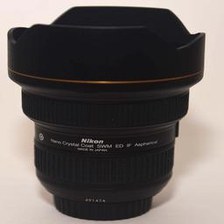}    & \includegraphics[width=.055\textwidth,height=.055\textwidth]{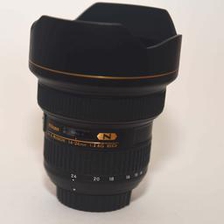}    & \includegraphics[width=.055\textwidth,height=.055\textwidth,cfbox=cyan]{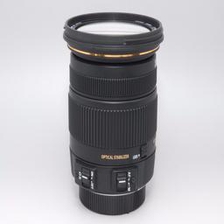}    & \includegraphics[width=.055\textwidth,height=.055\textwidth,cfbox=cyan]{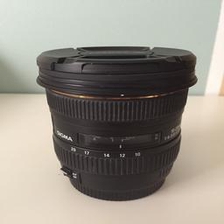}
    & ~~
    & 
	& \rotatebox[origin=l]{90}{\scalebox{0.5}{FG$_{Sum}$}}
    & \includegraphics[height=.055\textwidth]{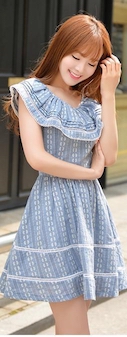}
    & \includegraphics[height=.055\textwidth,cfbox=cyan]{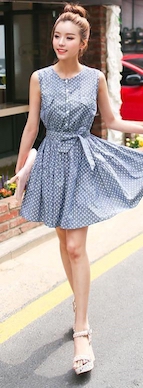}    
    & \includegraphics[height=.055\textwidth,cfbox=cyan]{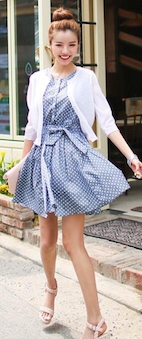}    
    & \includegraphics[height=.055\textwidth,cfbox=cyan]{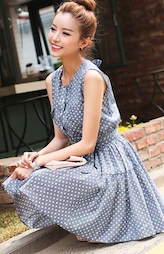}    
    & \includegraphics[height=.055\textwidth]{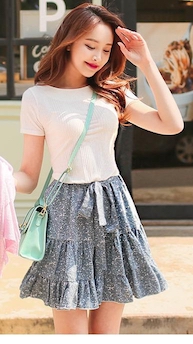}\\

    &
	\rotatebox[origin=l]{90}{\scalebox{0.5}{FG$_{Crow}$}}
    & \includegraphics[width=.055\textwidth,height=.055\textwidth,cfbox=cyan]{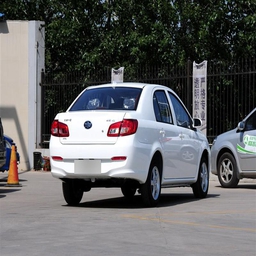}
    & \includegraphics[width=.055\textwidth,height=.055\textwidth,cfbox=cyan]{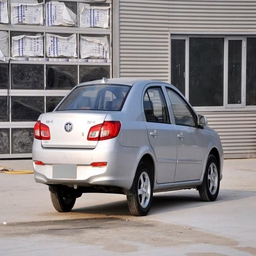}    & \includegraphics[width=.055\textwidth,height=.055\textwidth,cfbox=cyan]{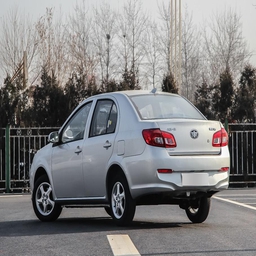}    & \includegraphics[width=.055\textwidth,height=.055\textwidth]{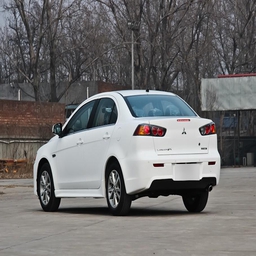}    & \includegraphics[width=.055\textwidth,height=.055\textwidth,cfbox=cyan]{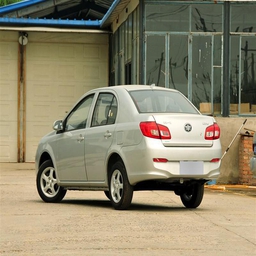}
    & ~~
    &
	& \rotatebox[origin=l]{90}{\scalebox{0.5}{FG$_{Crow}$}}
    & \includegraphics[width=.055\textwidth,height=.055\textwidth,cfbox=cyan]{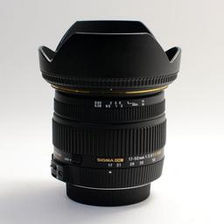}    & \includegraphics[width=.055\textwidth,height=.055\textwidth]{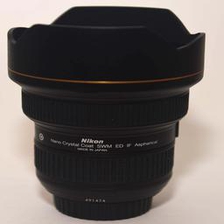}    & \includegraphics[width=.055\textwidth,height=.055\textwidth]{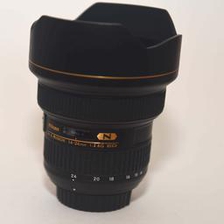}    & \includegraphics[width=.055\textwidth,height=.055\textwidth,cfbox=cyan]{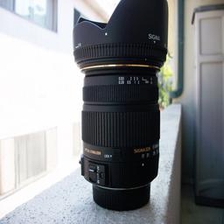}    & \includegraphics[width=.055\textwidth,height=.055\textwidth,cfbox=cyan]{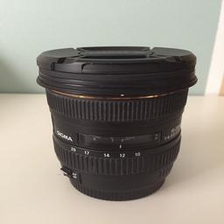}
    & ~~
    & 
	& \rotatebox[origin=l]{90}{\scalebox{0.5}{FG$_{Crow}$}}
    & \includegraphics[height=.055\textwidth,cfbox=cyan]
{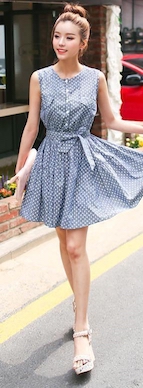}
    & \includegraphics[height=.055\textwidth]{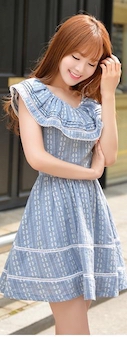}    
    & \includegraphics[height=.055\textwidth,cfbox=cyan]{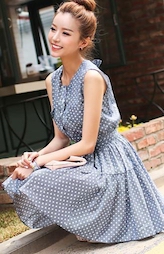}    
    & \includegraphics[height=.055\textwidth,cfbox=cyan]{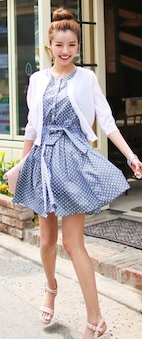}    
    & \includegraphics[height=.055\textwidth]
{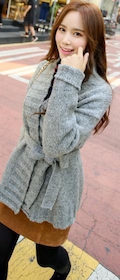}\\

    &
	\rotatebox[origin=l]{90}{\scalebox{0.5}{FG$_{SCDA}$}}
    & \includegraphics[width=.055\textwidth,height=.055\textwidth,cfbox=cyan]{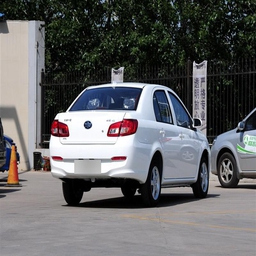}
    & \includegraphics[width=.055\textwidth,height=.055\textwidth]{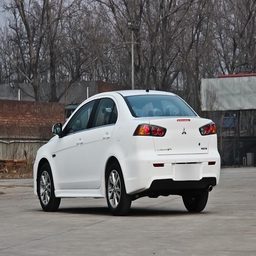}    & \includegraphics[width=.055\textwidth,height=.055\textwidth,cfbox=cyan]{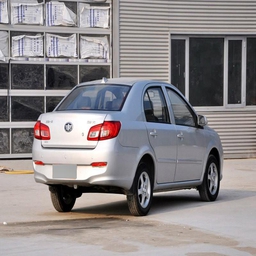}    & \includegraphics[width=.055\textwidth,height=.055\textwidth,cfbox=cyan]{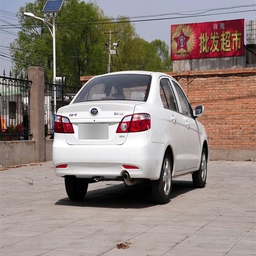}    & \includegraphics[width=.055\textwidth,height=.055\textwidth]{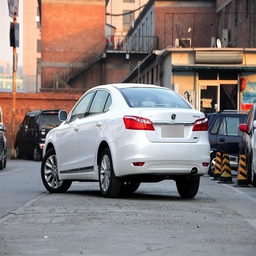}
    & ~~
    &
	& \rotatebox[origin=l]{90}{\scalebox{0.5}{FG$_{SCDA}$}}
    & \includegraphics[width=.055\textwidth,height=.055\textwidth,cfbox=cyan]{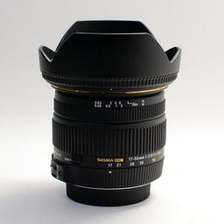}    & \includegraphics[width=.055\textwidth,height=.055\textwidth]{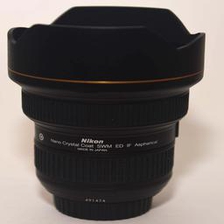}    & \includegraphics[width=.055\textwidth,height=.055\textwidth,cfbox=cyan]{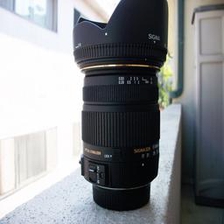}    & \includegraphics[width=.055\textwidth,height=.055\textwidth]{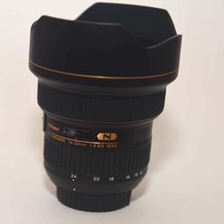}    & \includegraphics[width=.055\textwidth,height=.055\textwidth,cfbox=cyan]{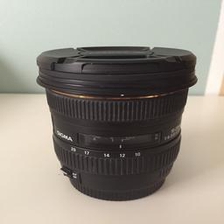}
    & ~~
    & 
	& \rotatebox[origin=l]{90}{\scalebox{0.5}{FG$_{SCDA}$}}
    & \includegraphics[height=.055\textwidth,cfbox=cyan]
{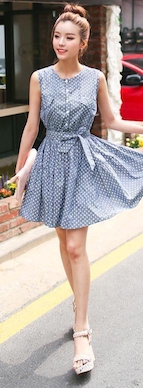}
    & \includegraphics[height=.055\textwidth]
{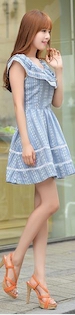}    
    & \includegraphics[height=.055\textwidth]{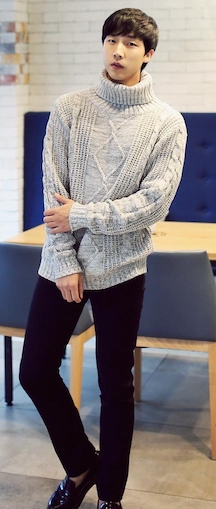}    
    & \includegraphics[height=.055\textwidth]{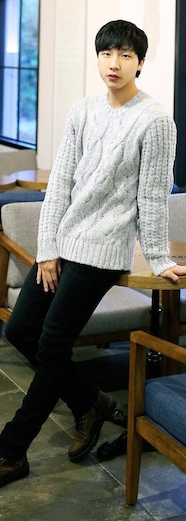}    
    & \includegraphics[height=.055\textwidth]
{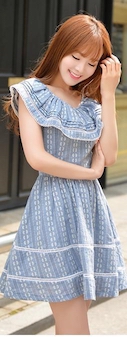}\\
    
 \end{tabular}
 \caption{Top 5 retrieved images from different methods on a sample query from CompCars, eBayCamera10k and Lookbook datasets. Images with blue border are true positive samples.}
\label{fig:query-retrieved-images-comcars} 
\end{figure*}

\subsection{Baselines comparisons}
\label{sec:baselines}
{We compare our FGGAN with off-the-shelf VGG16~\cite{Simonyan14c}, VGG16 fine-tuned on the target dataset (VGG16$+$\emph{ft}), several recent state-of-the-arts including MAC~\cite{tolias2015particular,azizpour2015generic}, Sum pooling~\cite{babenko2015aggregating}, CroW~\cite{kalantidis2016cross}, and SCDA~\cite{DBLP:journals/tip/WeiLWZ17} that are designed for fine-grained and generic image retrieval. 
These approaches aggregate the activations from the last pooling layer to form a compact representation for retrieval. For fair comparison, we apply these aggregation methods to our semantic embedding module $S$ instead of the plain VGG16$+$\emph{ft} network to show the advantages of our adversarial learning scheme.}
Table~\ref{tbl:map-compare} shows the performance comparison on three datasets in terms of mAP. 
The proposed FGGAN, using activations from the last fully-connected layer~\emph{fc} as features for search, achieves comparable or better performance than the state-of-the-art methods on the three datasets. 
By aggregating activations from the last pooling layer, our FGGAN still outperforms most of its rivals, verifying that our trained semantic embedding module is more effective for image search.
Again, our proposed network achieves higher or comparable accuracy than the compared methods. Qualitative results of retrieved images given a query are shown in Figure~\ref{fig:query-retrieved-images-comcars}.
Specifically, images from the CompCars and eBayCamera10k datasets only contain rigid objects, where the composition of cars and cameras across different views are usually consistent. 
In contrast, the Lookbook dataset consists of clothing images that are highly non-rigid and deformable under different viewpoints. Therefore, it is challenging to learn geometric transformations universally good for all non-rigid objects. {Although our method performs less favorably than the states-of-the-arts on Lookbook dataset, our FGGAN+Crow still achieve the best mAP at rank 5, showing its effectiveness in discovering correct images at an early stage.} 






\begin{figure*}[t]
	\centering
        \includegraphics[width=1.03\textwidth]{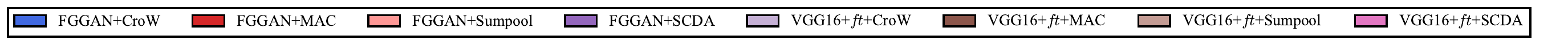}\\
	\subfloat{\label{fig:unseen-compcars-p-at-k}{\includegraphics[height=.2\textwidth]{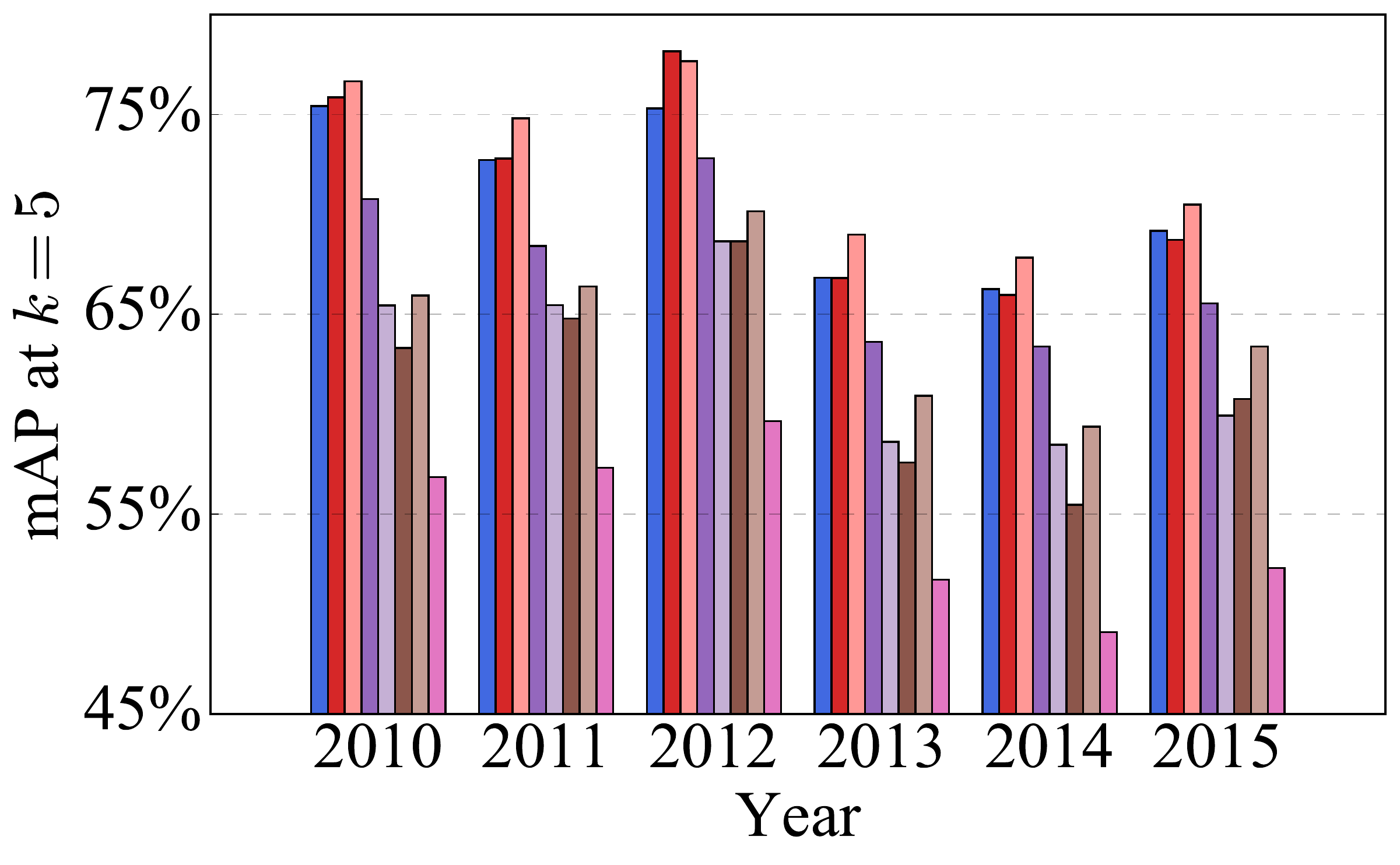}}} 
	\subfloat{\label{fig:unseen-compcars-p-at-k}{\includegraphics[height=.2\textwidth]{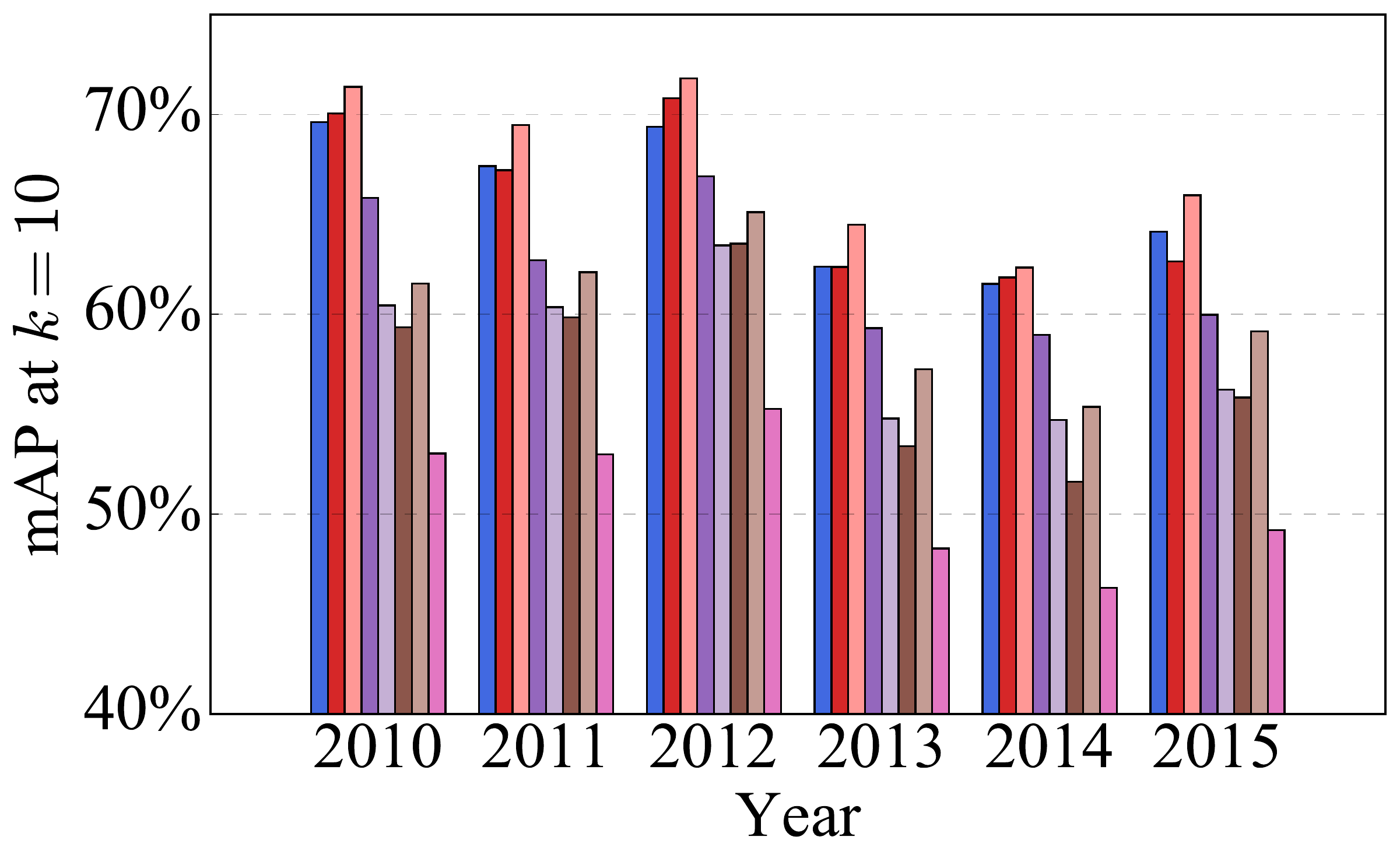}}} 
	\subfloat{\label{fig:unseen-compcars-p-at-k}{\includegraphics[height=.2\textwidth]{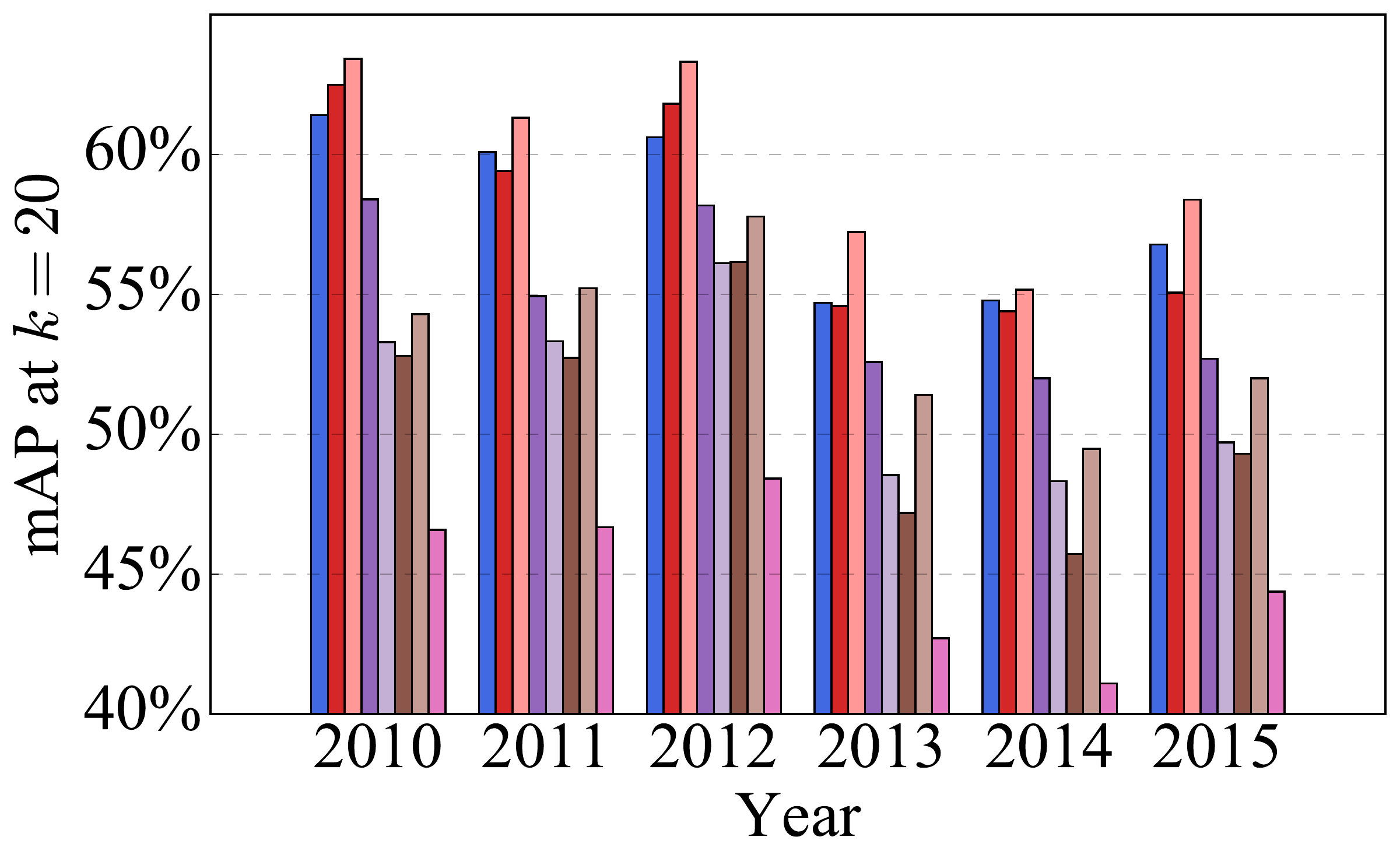}}} 
	\caption{mAP (\%) of different methods in the open-set scenario on the CompCars dataset. FGGAN outperforms the states-of-the-art throughout the years even if the database keeps growing in terms of size and contents. Note that all the compared models are trained on the cars appeared before the year of $2010$.
}
	\label{fig:compcars-unseen-p-at-k}
\end{figure*}

\subsection{Unseen image retrieval}
\label{sec:unseen-exp}
In the open-set scenario, initial training set is incomplete while new images and categories accumulate continuously.
We evaluate the generalization ability of our model trained on a subset of the entire dataset by measuring how well it adapts to new data.
We conduct experiments on the CompCars dataset as it provides rich annotations including the year of car manufacture.
Specifically, we train our network only on cars manufactured before the year of $2010$, and then test on new cars from the following years.
In each year, we take new test images as queries, and retrieve relevant images from the database that contains all known and unseen images so far. 
Figure~\ref{fig:compcars-unseen-p-at-k} shows the retrieval precision and mAP of different methods evaluated on top $k$ retrieved samples in the open-set scenario. FGGAN consistently outperforms the state-of-the-art approaches throughout all years even if the database keeps growing in terms of size and content.
Specifically, FGGAN$+$Sumpool achieve the best performance and improves baselines by a substantial margin.
The results suggest that, given a small amount of training data, our network can recognize and retrieve relevant images from unseen categories and is more generalizable.

\begin{table}[t]
\renewcommand{\arraystretch}{0.95}
	\centering
    \footnotesize
	\caption{Performance comparison (mAP, \%) for the open-set scenario with images from unseen categories in eBayCamera10k and Lookbook. Lookbook has nonrigid objects which are challenging for FGGAN. Section \ref{sec:unseen-exp} has details.}
	\label{tbl:unseen}
	\begin{tabular}{@{}@{\extracolsep{\fill}}l@{\hspace{2mm}}c@{\hspace{1mm}}c@{\hspace{1mm}}c@{}c|c@{\hspace{1mm}}c@{\hspace{1mm}}c@{}}
    & \multicolumn{3}{c}{eBayCamera10k} && \multicolumn{3}{c}{Lookbook Dataset} \\
	\hline
	Method  & $k = 5$ &  $k = 10$ & $k = 20$ && $k = 5$ &  $k = 10$ & $k = 20$\\
	\hline
    VGG16~\cite{Simonyan14c} & $78.66$ & $75.21$ & $72.17$ && $52.52$ & $50.52$ & $46.93$\\
    VGG16+\emph{ft}~\cite{Simonyan14c} & $82.53$ & $79.29$ & $76.19$ && $77.29$ & $73.37$ & $68.99$ \\
    \hline
 MAC+\emph{ft}~\cite{tolias2015particular,azizpour2015generic} & $83.53$ & $80.09$ & $76.98$ && $81.02$ & $76.46$ & $72.06$\\ 
    Sum pooling+\emph{ft}~\cite{babenko2015aggregating} & $82.89$ & $79.53$ & $76.30$ && $81.36$ & $77.22$ & $73.05$\\ 
    CroW+\emph{ft}~\cite{kalantidis2016cross} & $83.25$ & $80.03$ & $77.06$ && $\textbf{82.32}$ & $\textbf{77.98}$ & $\textbf{73.53}$\\ 
 SCDA+\emph{ft}~\cite{DBLP:journals/tip/WeiLWZ17} & $83.88$ & $80.61$ & $77.51$ && $67.60$ & $63.92$ & $59.50$\\  
    \hline
    FGGAN & $88.37$ & $85.92$ & $83.66$ && $70.79$ & $66.60$ & $61.19$\\ 
    FGGAN+MAC & $90.38$ & $88.27$ & $85.93$ && $73.48$ & $69.70$ & $65.39$\\ 
    FGGAN+Sum pooling & $90.21$ & $87.85$ & $85.18$ && $75.29$ & $70.92$ & $66.81$\\ 
    FGGAN+CroW & $\textbf{90.45}$ & $\textbf{88.40}$ & $\textbf{86.17}$ && $77.14$ & $72.63$ & $68.08$\\ 
    FGGAN+SCDA & $89.87$ & $87.97$ & $85.64$ && $60.34$ & $57.36$ & $53.84$\\ 
	\hline
	\end{tabular}
	\caption{Performance comparison (mAP, \%) of different module combinations.
    $N^{\star}$ denotes the normalizer without feature reconstruction.}
	\label{tbl:map-compare2}
    {
	\begin{tabular*}{\textwidth}{@{}@{\extracolsep{\fill}}lcccc|cccc|ccc@{}}
	& \multicolumn{3}{c}{CompCars} && \multicolumn{3}{c}{eBayCamera10k} && \multicolumn{3}{c}{Lookbook} \\
	\hline
	Method  & $k = 5$ &  $k = 10$ & $k = 20$ && $k = 5$ &  $k = 10$ & $k = 20$ && $k = 5$ &  $k = 10$ & $k = 20$\\
	\hline
    CAE ($G$)~\cite{masci2011stacked} & $14.96$ & $15.36$ & $14.92$ && $75.19$ & $70.79$ & $67.27$ && $5.96$ & $6.13$ & $6.20$\\
    DCGAN ($GD$)~\cite{radford2015unsupervised} & $15.74$ & $15.93$ & $15.29$ && $73.16$ & $67.68$ & $63.38$ && $6.87$ & $7.06$ & $7.13$\\
    \hline
    FGGAN $GDN^{\star}$ & $16.19$ & $16.56$ & $16.35$ && $72.22$ & $67.31$ & $63.39$ && $6.15$ & $6.48$ & $6.52$\\ 
    FGGAN $GDN$ & $16.52$ & $16.84$ & $16.35$ && $72.76$ & $67.71$ & $63.49$ && $8.74$ & $8.98$ & $8.73$\\
    FGGAN $GDN^{\star}S$ & $\textbf{66.60}$ & $62.19$ & $55.59$ && $\textbf{92.86}$ & $\textbf{91.01}$ & $\textbf{88.85}$ && $75.37$ & $71.21$ & $66.31$\\ 
    FGGAN $GDNS$ & $66.47$ & $\textbf{64.52}$ & $\textbf{56.28}$ && ${92.16}$ & ${90.34}$ & ${87.81}$ && $\textbf{77.00}$ & $\textbf{71.90}$ & $\textbf{66.98}$\\ 
	\hline
	\end{tabular*}
    }
\end{table}

Due to lack of year information in the eBayCamera10k dataset, we randomly split the training set into $50$ known camera categories and $60$ unseen camera categories to mimic the open-set scenario.
We train our model on the incomplete training set and test our model on the original test set consisting of $50$ known and $60$ unseen camera categories. 
In Table~\ref{tbl:unseen}, we observe that our method performs more favorably against the states-of-the-arts even though the database contains a large number of unknown categories. Similarly, we split the Loobook dataset to $4,000$ known and $4,726$ unseen clothing categories, and report the open-set experiments in Table~\ref{tbl:unseen}. The results are consistent with previous finding that it is challenging to learn geometric transformations for non-rigid clothing images. 



\subsection{Ablation study}
\label{sec:ablation-study}

{\flushleft \textbf{Influence of individual modules}.}
We compare several variants of our FGGAN with DCGAN~\cite{radford2015unsupervised} and Convolutional AutoEncoder (CAE)~\cite{masci2011stacked}. 
While DCGAN consists of a generator $G$ and a discriminator $D$, the CAE can be seen as a single generator $G$. 
{All comparisons are using the raw features from the last fully-connected layer while we observe similar results when using aggregated features.}
As shown in Table~\ref{tbl:map-compare2}, FGGAN $GDN$ performs more favorably against DCGAN ($GD$) and CAE ($G$) on the CompCars and Lookbook datasets. 
The results indicate that converting images in various views to the canonical view reduces view ambiguity and is useful for image retrieval. 
In addition, FGGAN $GDN$ achieves higher mAP than $GDN^\star$, where $N^\star$ denotes the normalizer without the feature reconstruction loss. 
The results suggest that the feature reconstruction loss is helpful for learning effective representations as the regularization term enforces the generated representations to be close to the canonical images in feature space. 
FGGAN $GDNS$ and FGGAN $GDN^{\star}S$ achieve comparable results and outperform all other compared models, which shows that including the semantic embedding is critical for learning discriminative features.
On the eBayCamera10k dataset, since different categories of camera may have visually similar stock images, it is challenging to learn an effective manifold by plain GAN. 
Therefore, we notice that FGGAN $GDNS$, which learns with semantic category supervision, infers the correct manifold more effectively.

{\flushleft \textbf{Different backbone network}.} To demonstrate that our FGGAN is a generic meta-architecture applicable to any backbone networks, we replace VGG16 by ResNet101 for the semantic embedding module and evaluate its performance. We compare our FGGAN with a recent image retrieval approach~\cite{gordo2017end} using ResNet101 as a backbone network with RPN and triplet loss. Without any additional components and expensive annotations such as triplets, our FGGAN clearly surpasses~\cite{gordo2017end} on all datasets, indicating that our adversarial learning scheme is more effective in producing discriminative features to distinguish fine-grained categories, while being more flexible. 

{\flushleft \textbf{Different views}.} In Section~\ref{sec:baselines}, we present the results on the CompCars dataset using front view as the canonical view. 
It is worth noting that the proposed FGGAN does not set any restriction on the canonical view. 
We experiment with other views as canonical views and evaluate the corresponding performance. 
In Table~\ref{tbl:multi-view}, using the side views leads to the best mAP when $k=5$, meaning that it is more effective in finding correct matches at an early stage.
Since there are more images in the side view than those in the front view, they provide more diverse samples for training to improve the network's generalizability.
Overall, the mAP is consistent across all views with negligible difference, which clearly shows that FGGAN is robust in discovering the underlying manifold and finding correct matches from various views. 

\begin{table*}[ht]
\renewcommand{\arraystretch}{0.95}
	\centering
    \small
    	\caption{Performance comparison (mAP, \%) of using ResNet101 as the backbone network.}
	\label{tbl:map-compare-resnet}
    \vspace{-2mm}
    {
	\begin{tabular*}{\textwidth}{@{}@{\extracolsep{\fill}}lcccc|cccc|ccc@{}}
	& \multicolumn{3}{c}{CompCars} && \multicolumn{3}{c}{eBayCamera10k} && \multicolumn{3}{c}{Lookbook} \\
	\hline
	Method  & $k = 5$ &  $k = 10$ & $k = 20$ && $k = 5$ &  $k = 10$ & $k = 20$ && $k = 5$ &  $k = 10$ & $k = 20$\\ 
    \hline
    ResNet101~\cite{he2016deep} & $59.54$ & $55.87$ & $49.82$ && $83.61$ & $80.82$ & $77.85$ && $65.74$ & $62.27$ & $57.80$\\
    RPN$+$TripletLoss~\cite{gordo2017end} & $78.65$ & $73.32$ & $65.52$ && $86.44$ & $83.30$ & $78.88$ && $71.89$ & $68.96$ & $65.58$\\
	FGGAN$_{ResNet101}$& $\textbf{81.99}$ & $\textbf{76.65}$ & $\textbf{68.83}$ && $\textbf{95.68}$ & $\textbf{94.89}$ & $\textbf{94.06}$ && $\textbf{87.68}$ & $\textbf{83.02}$ & $\textbf{78.63}$\\
	\hline
	\end{tabular*}
    }
%
	\caption{Performance comparison (mAP, \%) of FGGAN evaluated on top $k$ retrieved images when trained with different canonical views. Mean is the average over mAP values at $k$ = 5, 10 and 20. Details in Section \ref{sec:ablation-study}.}
	\label{tbl:multi-view}
	\begin{tabular*}{\columnwidth}{@{}@{\extracolsep{\fill}}lcccc@{}}
	\hline
	Canonical view  & $k = 5$ &  $k = 10$ & $k = 20$ & Mean\\
	\hline
    rear-side & $64.82$ & $60.97$ & $54.22$ & $60.00$\\
    rear & $65.64$ & $61.16$ & $54.58$ & $60.46$\\
    side & $\textbf{66.89}$ & $61.80$ & $55.07$ & $61.25$\\
    front-side & $66.82$ & $62.20$ & $56.18$ & $61.73$\\
    front & $66.47$ & $\textbf{62.52}$ & $\textbf{56.28}$ & $\textbf{61.75}$\\
	\hline
	\end{tabular*}
	\caption{Performance comparison (mAP, \%) of the proposed method evaluated on top $k$ retrieved images with the feature reconstruction loss applied to different layers. Mean is the average over the mAP values at $k=5$, $10$ and $20$.}
	\label{tbl:feat-recon}
	\begin{tabular*}{\columnwidth}{@{}@{\extracolsep{\fill}}lcccc@{}}
	\hline
	Convolutional layer  & $k = 5$ &  $k = 10$ & $k = 20$ & Mean\\
	\hline
    \emph{conv1} & $66.52$ & $62.27$ & $55.33$ & $61.37$\\
    \emph{conv2} & $\textbf{66.66}$ & $62.32$ & $55.57$ & $61.51$\\
    \emph{conv3} & $65.82$ & $61.71$ & $54.70$ & $60.74$\\
    \emph{conv4} & $66.47$ & $\textbf{62.52}$ & $\textbf{56.28}$ & $\textbf{61.75}$\\
	\hline
	\end{tabular*}
\end{table*}

{\flushleft \textbf{Different feature reconstruction losses}.}
In Eq.~\eqref{eq:n}, the feature reconstruction loss enforces that the generated representations are semantically similar to the images in the canonical view in terms of the features extracted from the convolutional layer \emph{conv4} of the normalizer $N$. 
Since the loss is not limited to a specific convolutional layer, we attach it to different convolutional layers and show the performance in Table~\ref{tbl:feat-recon}. 
While using lower-level information (\emph{conv2}) is more advantageous for $k=5$, the higher layer (\emph{conv4}) is also important as it encodes more high-level semantics suitable for fine-grained image search.

\begin{table}[ht]
\renewcommand{\arraystretch}{0.95}
	\caption{Multi-label based retrieval with stricter evaluation criterion. Performance comparison (mAP, \%) of the proposed method on CompCars and eBayCamera10k. Notice large improvements for eBayCamera10k since fine-grained difference is subtle.
    }
	\label{tbl:strict}
    \vspace{-2mm}
    \centering
    \footnotesize
	\begin{tabular}{@{}@{}l@{\hspace{2mm}}c@{\hspace{1mm}}c@{\hspace{1mm}}c@{}c|c@{\hspace{1mm}}c@{\hspace{1mm}}c@{}}
	& \multicolumn{3}{c}{CompCars} && \multicolumn{3}{c}{eBayCamera10k} \\
    \hline
	Method  & $k = 5$ &  $k = 10$ & $k = 20$ && $k = 5$ &  $k = 10$ & $k = 20$\\
	\hline
    VGG16~\cite{Simonyan14c} & $45.27$ & $43.76$ & $40.22$ && $28.49$ & $28.03$ & $25.76$\\
    VGG16+\emph{ft}~\cite{Simonyan14c} & $62.82$ & $59.03$ & $53.25$ && $39.94$ & $37.85$ & $33.89$\\
    \hline
 MAC+\emph{ft}~\cite{tolias2015particular,azizpour2015generic} & $\textbf{64.64}$ & $60.81$ & $55.54$ && $40.65$ & $38.42$ & $34.28$\\  
    Sum pooling+\emph{ft}~\cite{babenko2015aggregating} & $63.93$ & $60.79$ & $\textbf{55.81}$ && $38.35$ & $36.77$ & $33.03$\\  
    CroW+\emph{ft}~\cite{kalantidis2016cross} & $62.86$ & $59.54$ & $54.75$ && $40.67$ & $38.73$ & $34.35$\\  
 SCDA+\emph{ft}~\cite{DBLP:journals/tip/WeiLWZ17} & $62.17$ & $59.03$ & $53.98$ && $40.29$ & $38.23$ & $34.20$\\
    \hline
    FGGAN & $63.07$ & $59.31$ & $54.41$ && $55.75$ & $53.59$ & $\textbf{49.52}$\\ 
    FGGAN+MAC & ${64.53}$ & $\textbf{61.07}$ & ${55.56}$ && $56.40$ & $53.67$ & $48.60$\\ 
    FGGAN+Sum pooling & $64.35$ & $60.35$ & $54.53$ && $54.40$ & $51.55$ & $47.30$\\ 
    FGGAN+CroW & $63.81$ & $59.96$ & $54.64$ && $\textbf{56.80}$ & $\textbf{54.28}$ & $49.13$\\ 
    FGGAN+SCDA & $62.82$ & $59.67$ & $54.25$ && $56.13$ & $53.17$ & $48.57$\\ 
	\hline
	\end{tabular}
    \vspace{-2mm}
\end{table}

{\flushleft \textbf{Multi-label retrieval}.} In the real-world scenario, one may want to retrieve products not only of the same make but also of the same model as for query. 
We measure the relevance of retrieved images by checking whether they share exactly the same make and model as the query to evaluate FGGAN in this scenario. 
Since images from the Lookbook dataset do not have hierarchical labels, \eg, make and model, we conduct experiments on CompCars and eBayCamera10k datasets only. 
Table~\ref{tbl:strict} shows the performance comparison. FGGAN still improves precision of compared methods in most cases under stricter evaluation criterion. The observation indicates that the generated representation captures subtle visual differences and is discriminative to different fine-grained cars and cameras.

\begin{figure}[t]
\begin{center}
\includegraphics[width=.9\linewidth]{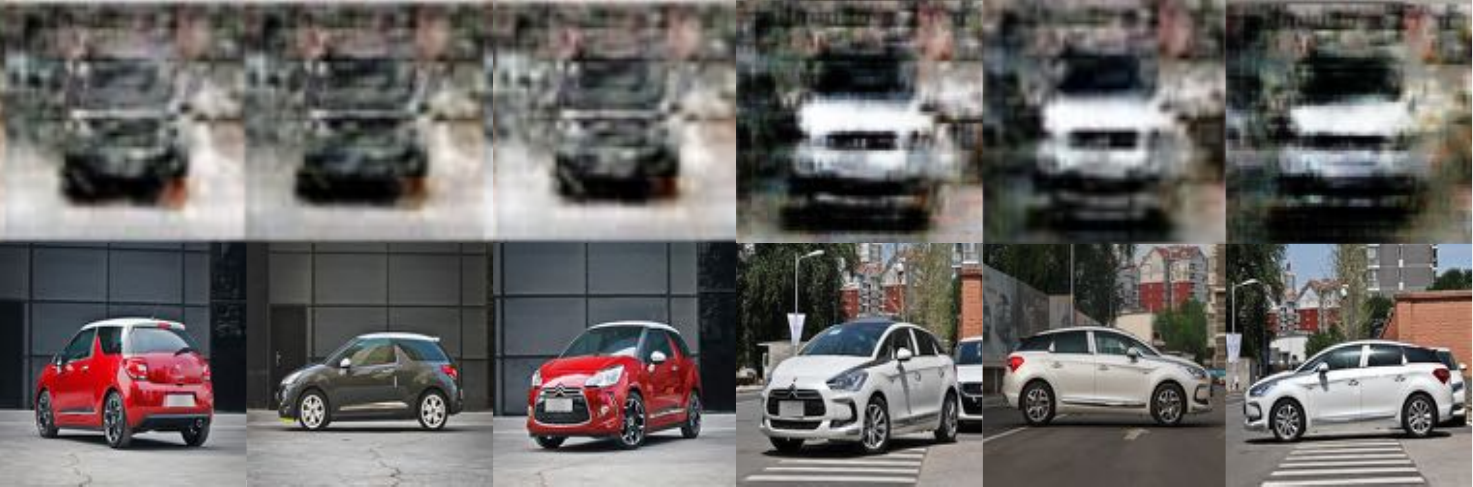}
\end{center}
\vspace{-4mm}
\caption{Visualization of the generated representations with front view as the canonical view. The left three cars are Citroen DS3, while the right three cars are Citroen DS5.}
	\label{fig:canonical view2}
\end{figure}

{\flushleft \textbf{Visualization of learned representations}.} We feed the real image $x_r$ into $G$, and visualize the generated representation $x_z$ in the RGB color space. In Figure~\ref{fig:canonical view2}, the images from the same category from the CompCars dataset are converted to similar representations even though they are in different views initially. 
We also notice that our generator is robust to the changes in color and illumination. 

%% file: 05-conclusion.tex
\section{Conclusion}\label{sec:conclusion}

We have presented an end-to-end network with adversarial learning for fine-grained image search. 
We have integrated a generative adversarial network (GAN) to learn implicit geometric transformations for view and pose normalization. 
Features extracted from our network is more discriminative to distinguish subtle differences of objects from fine-grained categories.
In an open-set scenario, our network is able to correctly match unseen images from unknown categories, given an incomplete training set, which is more scalable as data from new categories accumulates. 